\documentclass[runningheads]{llncs}
\usepackage[T1]{fontenc}
\usepackage{graphicx}
\usepackage{booktabs}
\usepackage{pifont}
\usepackage{amsfonts}
\newcommand{\cmark}{\ding{51}} 
\newcommand{\xmark}{\ding{55}} 

\usepackage{color}
\usepackage{multirow}
\usepackage{lipsum}
\def\our{GenPlanner}

\begin{document}
\title{GenPlanner: From Noise to Plans - Emergent Reasoning in Flow Matching and Diffusion Models}
\titlerunning{GenPlanner: From Noise to Plans - Emergent Reasoning in Diffusion Models}
%
\author{Agnieszka Polowczyk\inst{1}\orcidID{0009-0008-1583-4493} \and
Alicja Polowczyk\inst{1}\orcidID{0009-0001-3110-8255} \and
Michał Wieczorek\inst{1}\orcidID{0000-0002-5319-3366}}
\authorrunning{Polowczyk et al.}
%
\institute{Silesian University of Technology, Faculty of Applied Mathematics, Gliwice, 44-100, Poland}
%
\maketitle              
\begin{abstract}
Path planning in complex environments is one of the key problems of artificial intelligence because it requires simultaneous understanding of the geometry of space and the global structure of the problem. In this paper, we explore the potential of using generative models as planning and reasoning mechanisms. We propose GenPlanner, an approach based on diffusion models and flow matching, along with two variants: DiffPlanner and FlowPlanner. We demonstrate the application of generative models to find and generate correct paths in mazes. A multi-channel condition describing the structure of the environment, including an obstacle map and information about the starting and destination points, is used to condition trajectory generation. Unlike standard methods, our models generate trajectories iteratively, starting with random noise and gradually transforming it into a correct solution. Experiments conducted show that the proposed approach significantly outperforms the baseline CNN model. In particular, FlowPlanner demonstrates high performance even with a limited number of generation steps.

\end{abstract}
\section{Introduction}
Recent years have seen a rapid development of generative models and large language models (LLMs), which demonstrate significant capabilities in many domains. Despite these successes, tasks requiring logical reasoning and spatial planning still pose significant challenges. Current models struggle to solve problems like puzzles, which are intuitive to humans but require multi-step reasoning for machines \cite{chen2025enigmata}. This problem is particularly evident in high-dimensional navigation tasks, where the model's ability to find the correct solution drastically decreases. Standard Vision-Language Models (VLMs) often hallucinate, failing to maintain logical consistency over long planning horizons \cite{makinski2025reasoning}. There are comprehensive machine learning-based planning approaches that avoid sequential methods \cite{lin2025zebralogic}. Additionally, diffusion-based solutions have been proposed that utilize computationally expensive inference-time optimizations \cite{liang2025simultaneous}. In navigation tasks, global trajectory modeling using additional guidance mechanisms is crucial \cite{lu2025what}.

\begin{figure}[t]
\centering
\setlength{\tabcolsep}{1pt}
\renewcommand{\arraystretch}{0.8}
\begin{tabular}{ccccc}
 & 8 $\times$ 8 & 16 $\times$ 16 & 32 $\times$ 32 & 48 $\times$ 48\\
\rotatebox{90}{\hspace{9pt}{Ground Truth}} &
\includegraphics[width=0.23\textwidth]{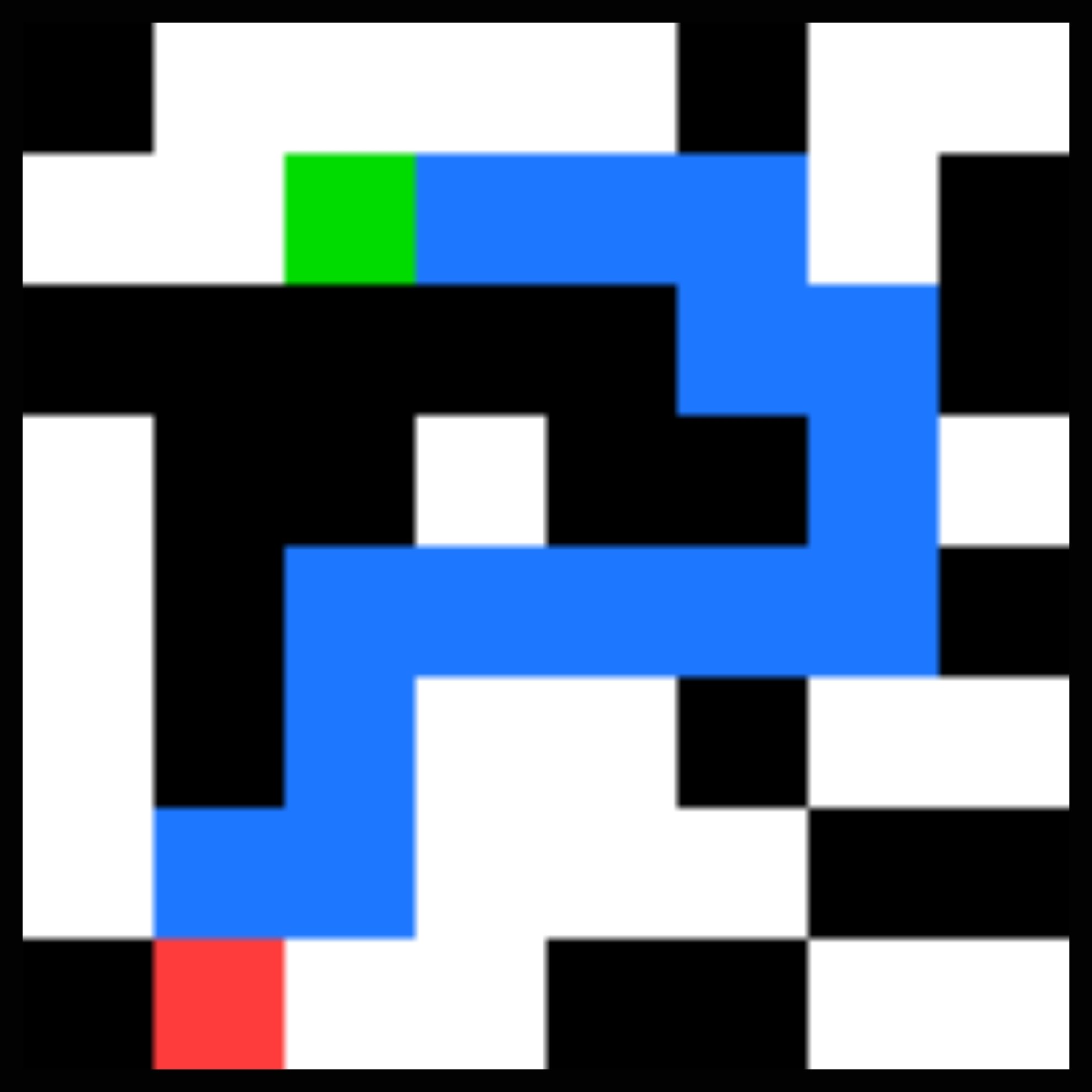} &
\includegraphics[width=0.23\textwidth]{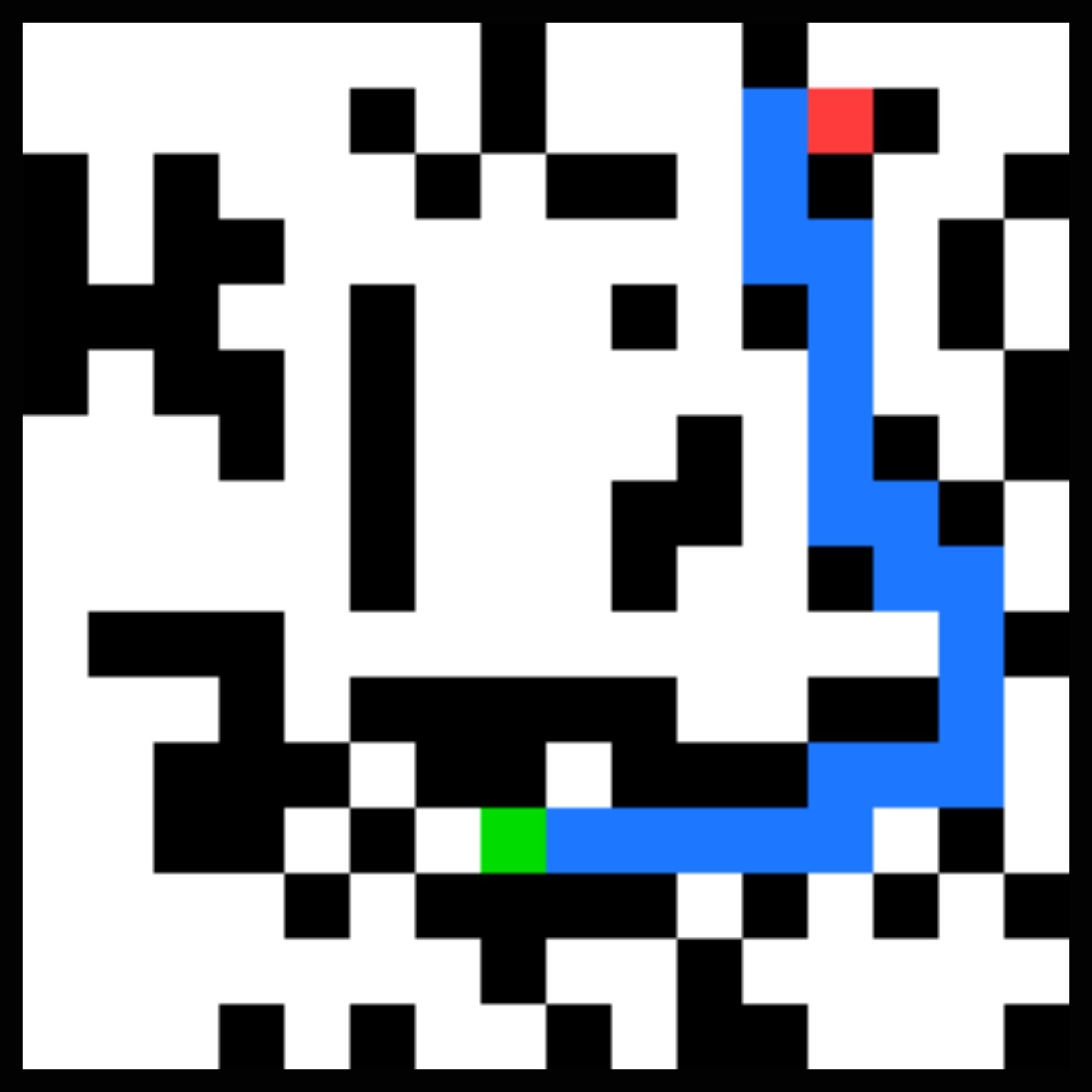} &
\includegraphics[width=0.23\textwidth]{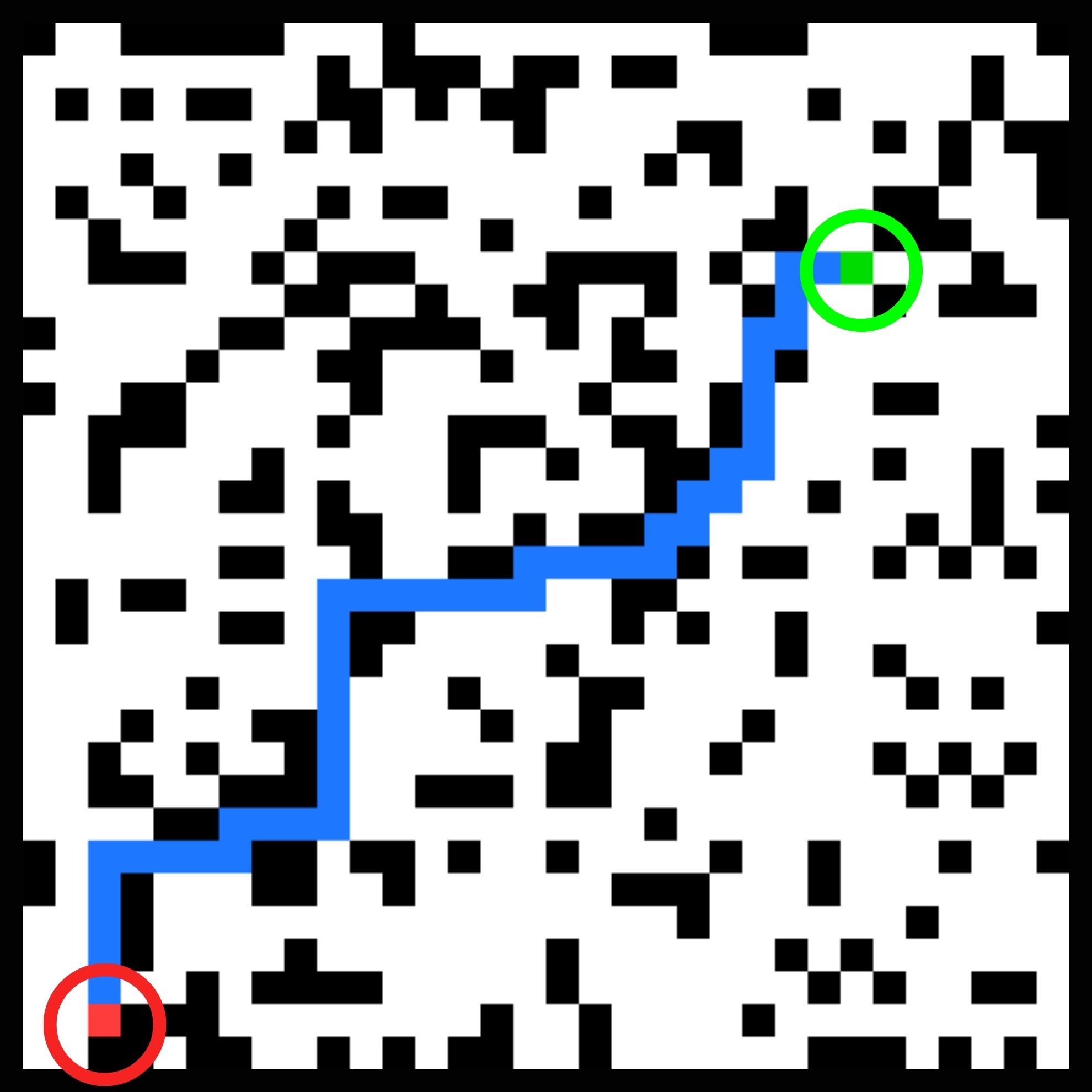} &
\includegraphics[width=0.23\textwidth]{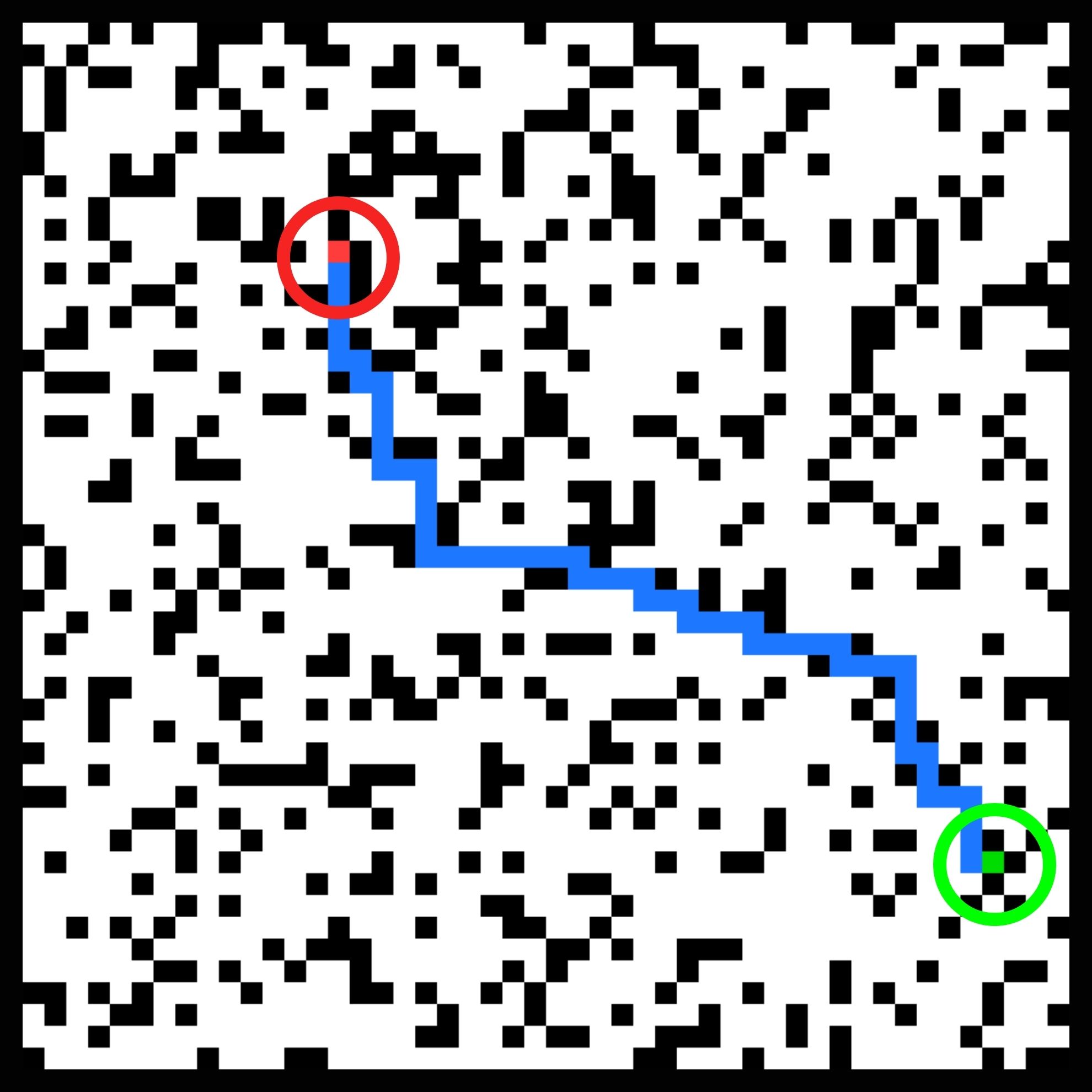} \\
\rotatebox{90}{\hspace{13pt}{FlowPlanner}} &
\includegraphics[width=0.23\textwidth]{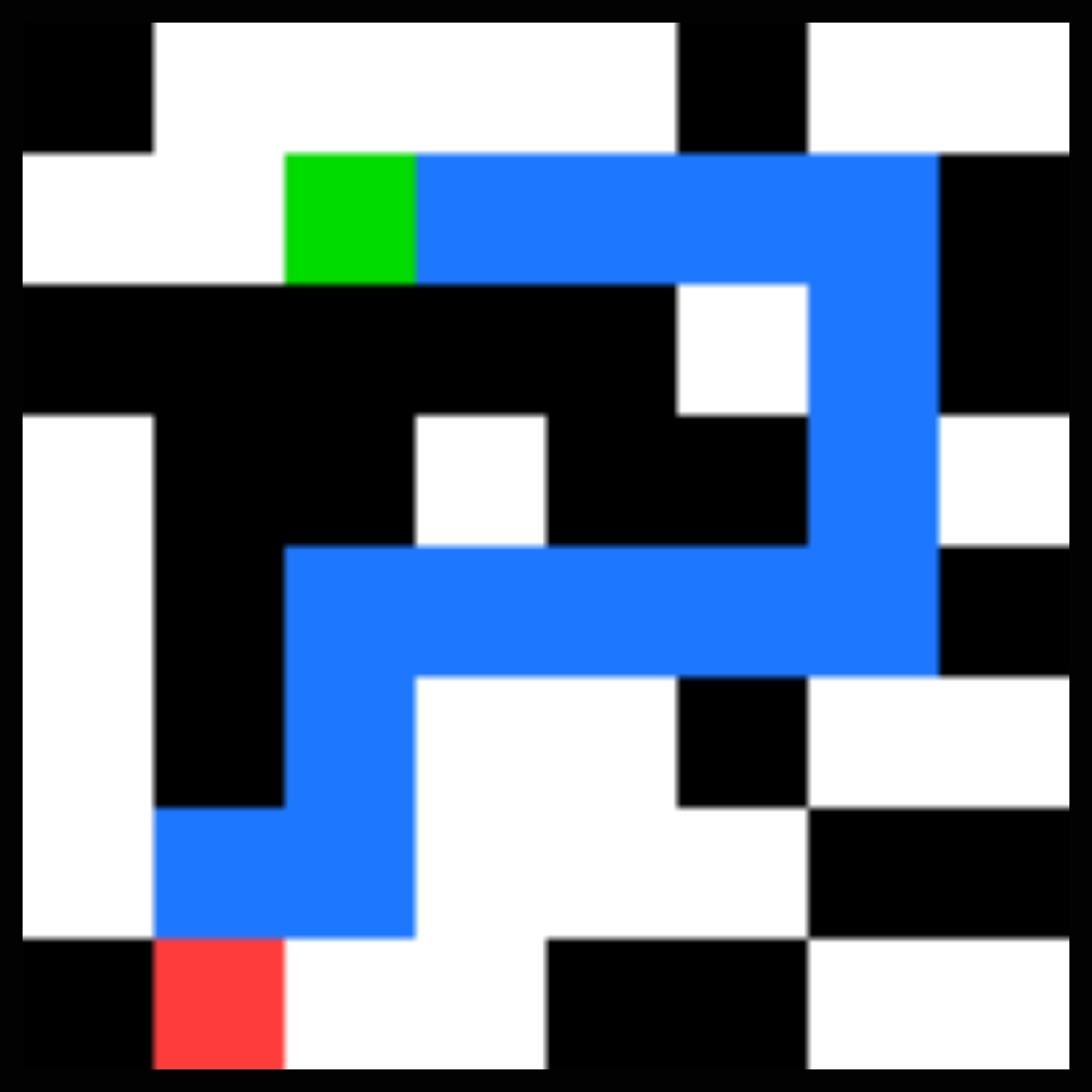} &
\includegraphics[width=0.23\textwidth]{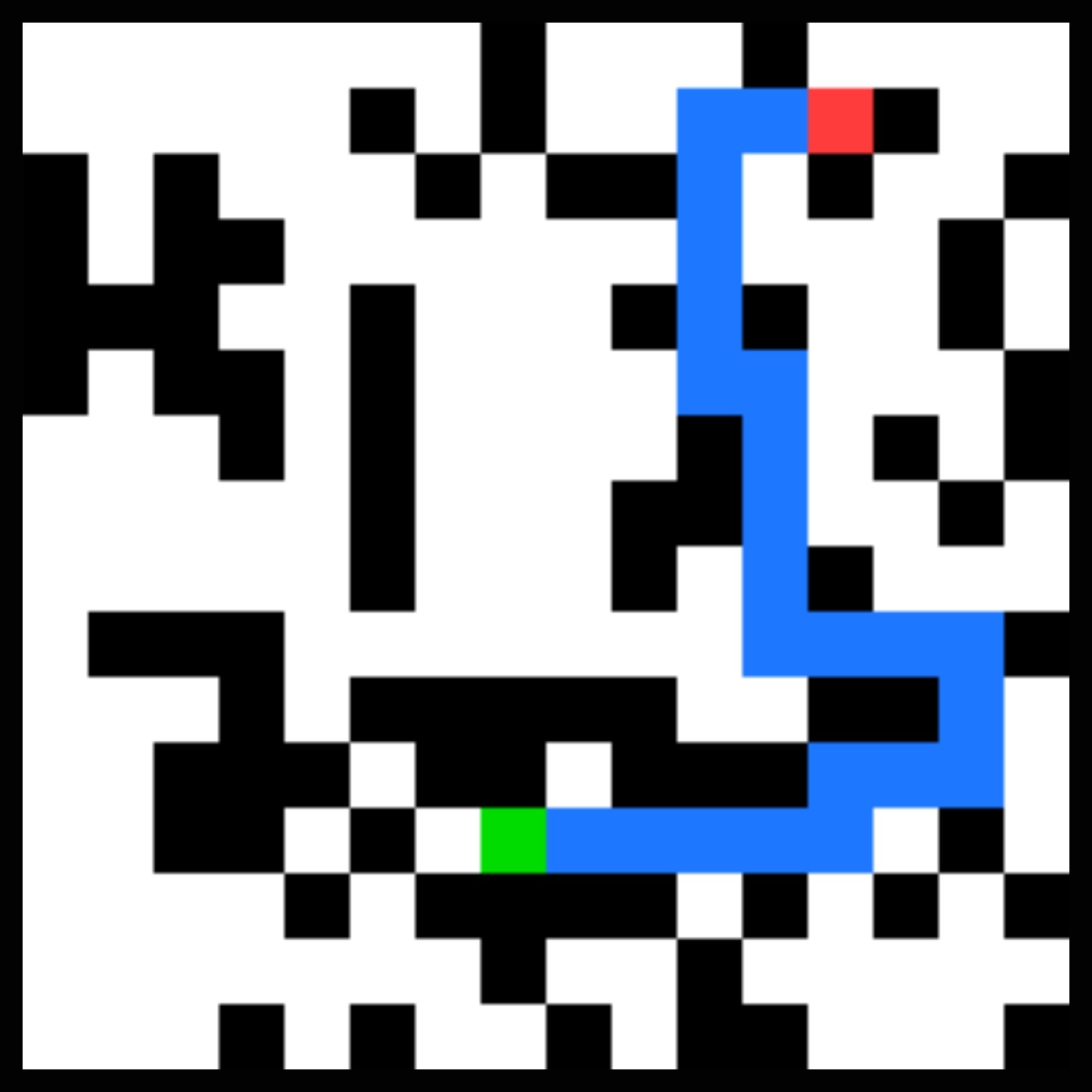} &
\includegraphics[width=0.23\textwidth]{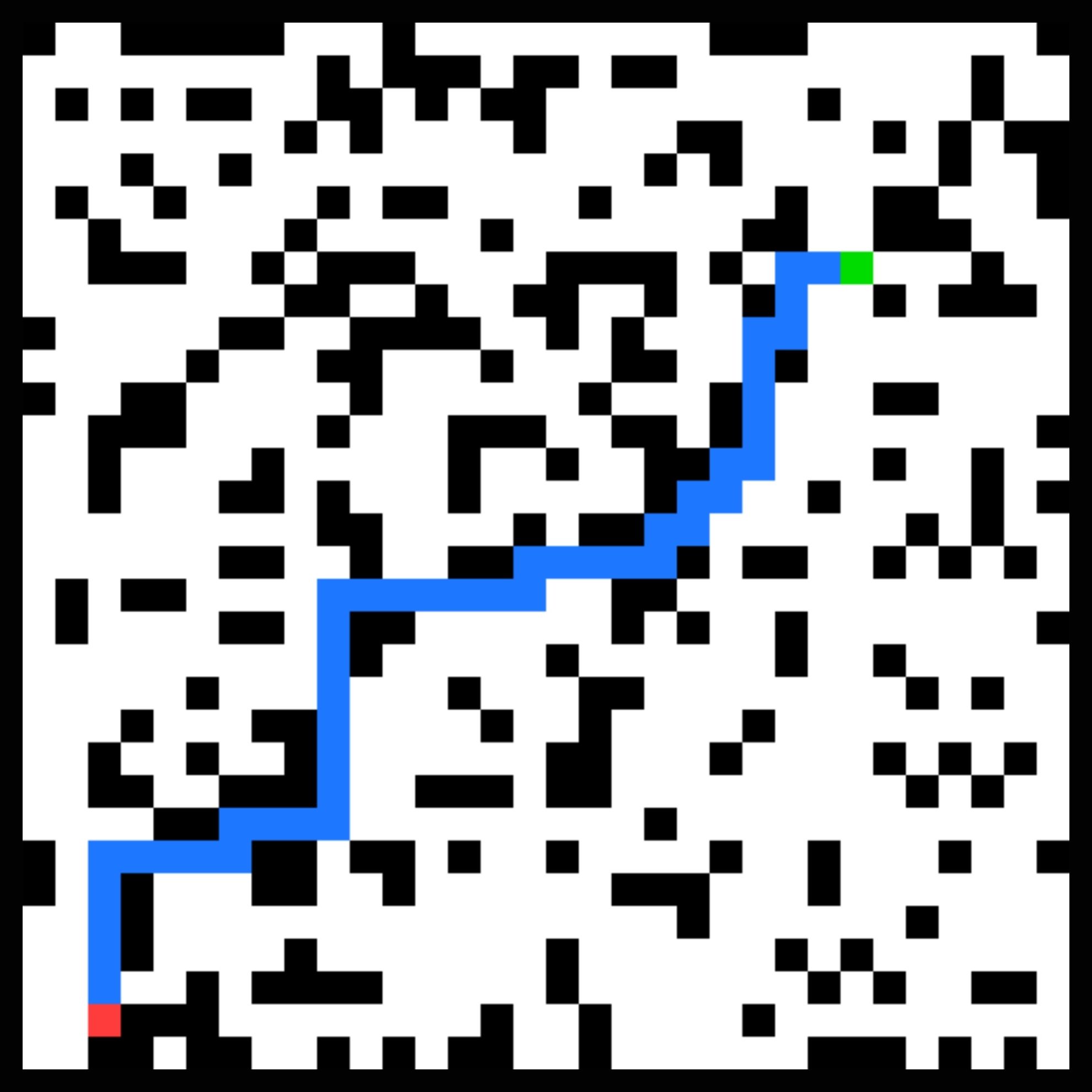} &
\includegraphics[width=0.23\textwidth]{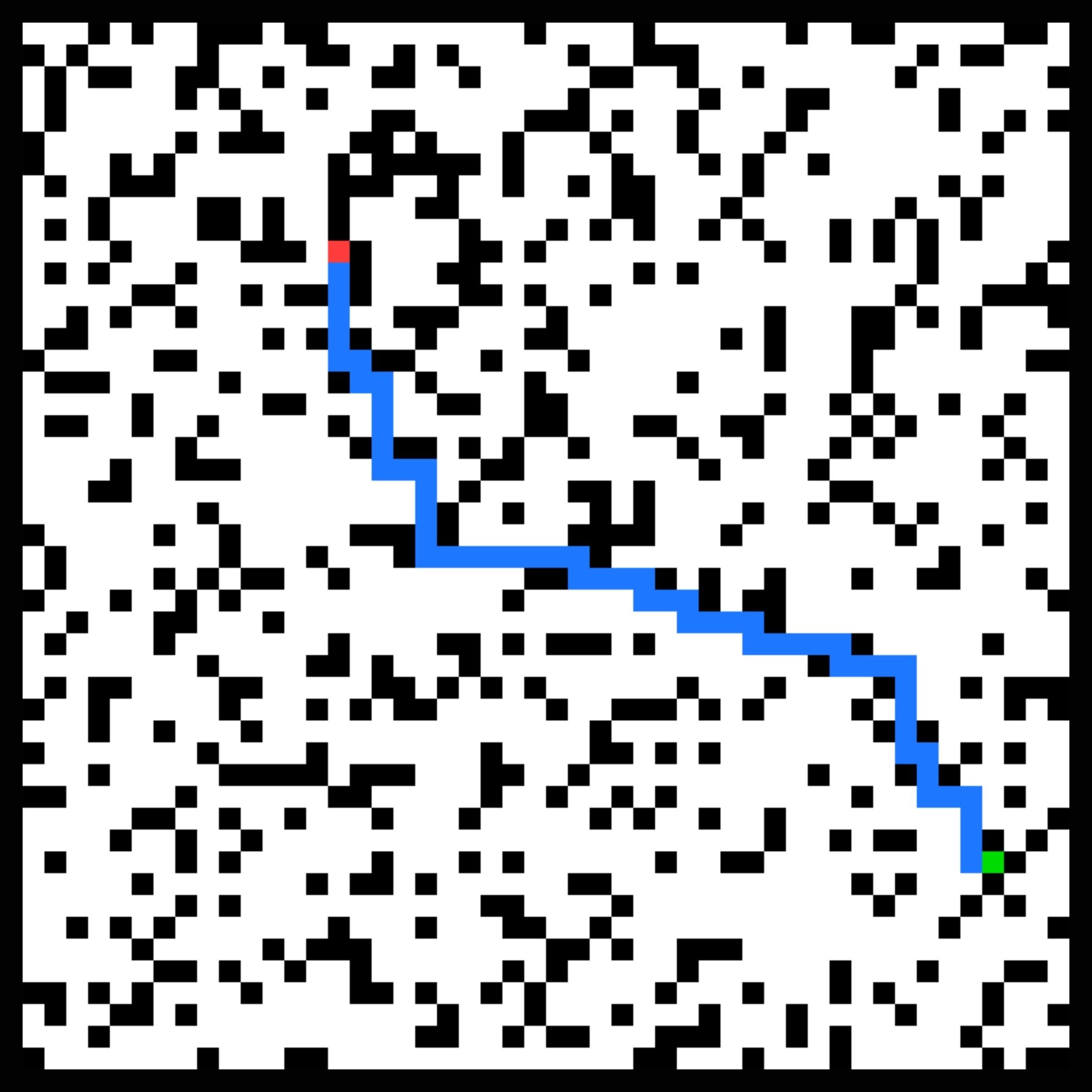} \\
\end{tabular}
\caption{\textbf{Qualitative comparison of ground-truth paths and FlowPlanner generations on grid mazes of different sizes ($8\times8$, $16 \times16$, $32\times32$, $48\times48$).} FlowPlanner generates trajectories consistent with reference paths, maintaining correctness and consistency even for larger mesh sizes.}
\label{fig: teaser}
\end{figure}

In this work, we propose a novel approach \our{} that solves the problem of planning in a maze with varying mesh sizes by formulating it as an image generation task, see Fig. \ref{fig: teaser}. Instead of using expensive optimization during inference, we developed the FlowPlanner and DiffPlanner algorithms based on generative models: Flow Matching and Diffusion, which allow for more stable and efficient training. Our model takes as input a noisy mesh and a conditioning tensor, whose individual channels encode the starting position, goal, and wall layout. This allows the model to implicitly learn the physics of the environment and geometric constraints. Experiments show that our method achieves a path generation efficiency of 89\% for solving 48 $\times$ 48 mazes, significantly outperforming approaches based on a baseline network that does not use iterative denoising.
In summary, our principal contributions are as follows:
\begin{itemize}
\item We introduce a new representation of the planning problem in the form of a multi-channel conditional tensor (start, end, and walls), which allows the generative model to learn obstacle avoidance.
\item We introduce GenPlanner in two variants: DiffPlanner and FlowPlanner, for planning and reasoning tasks.
\item We experimentally demonstrate that our model effectively handles the growing solution space for larger meshes of dimensions $48 \times 48$ with a high accuracy of 89\%, unlike the baseline CNN model, which fails in such conditions.
\end{itemize}

\section{Related Works}
Contemporary research confirms that, despite great advances in vision tasks, VLM models encounter fundamental difficulties in tasks requiring reasoning. These models struggle to abstract rules and perform worse than humans even on real-world images \cite{lin2025zebralogic}, \cite{makinski2025reasoning}, \cite{puzzlevqa}. Furthermore, benchmarks such as VisuLogic \cite{xu2026visulogic} and ENIGMATA \cite{chen2025enigmata} reveal that models often generate semantically correct but logically inconsistent solutions, as in the case of mazes, manifested by hallucinating paths through obstacles. Recent studies based on the DynaMath \cite{zou2025dynamath} and PARTNR \cite{chang2025partnr} benchmarks confirm VLM's sensitivity to small visual variances in mathematical tasks and its shortcomings in long-term spatial planning. Problems are also observed in tasks requiring lateral thinking and Boolean satisfiability problems \cite{wei2025satbench}, where the models exhibit so-called satisfiability bias. The application of diffusion models to planning is a rapidly developing area, employing global attention mechanisms \cite{lu2025what} or costly constrained optimization to enforce collision-free behavior \cite{liang2025simultaneous}. In the context of robotics, datasets such as RoboCerebra \cite{han2025robocerebra} and WOMB \cite{pmlr-v267-li25l} emphasize the need to integrate navigation with the physics of the environment. A key aspect of our algorithm is treating the planning problem as a holistic task, similar to the conclusions from the work \cite{unified}, which suggests that a unified image representation promotes better generalization. Unlike methods based on decomposition \cite{ryu2025divide}, our model solves the problem through a denoising process. Furthermore, we are also inspired by the observations from the work \cite{makinski2025reasoning}, which shows that longer reasoning correlates with correctness on difficult tasks. In our case, this role is played by an iterative diffusion process.

\section{Problem Formulation}
\begin{figure}[!t]
    \centering
    \includegraphics[width=1.0\linewidth]{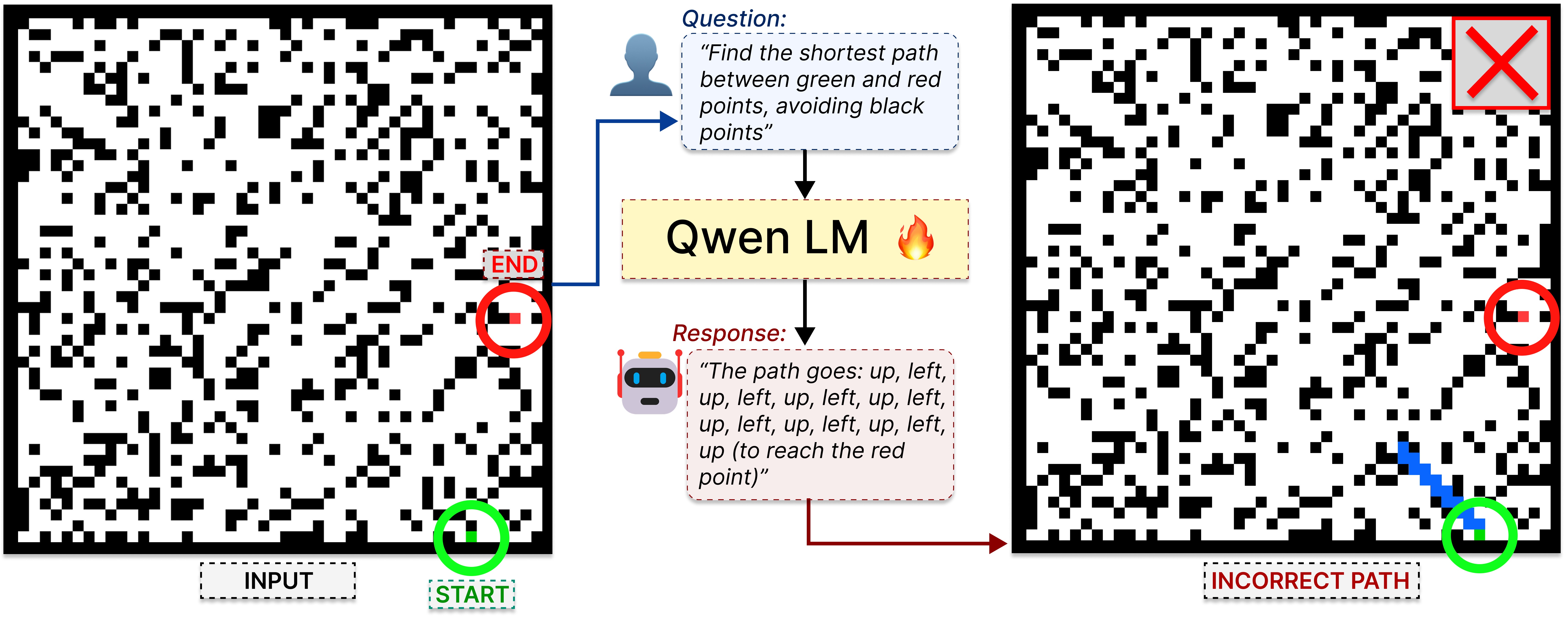}
    \caption{\textbf{Failure case of logical and spatial reasoning in a vision–language model.} Given a maze image with start and goal locations, the VLM (Qwen) generates a plausible sequence of directional moves, yet the resulting path is invalid. This example highlights the difficulty of current VLMs in performing precise logical reasoning and grid-based path planning.}
    \label{fig:vlm}
\end{figure}

Navigating complex, intricate 2D spaces, such as mazes poses significant challenges for current solutions to large-scale generative and linguistic models. This application introduces the need for long-range navigation and a space of logical constraints in the environments. Current research on solving these puzzles highlights the "curse of complexity" phenomenon \cite{lin2025zebralogic}, where the ability of models to find a correct solution decreases exponentially with increasing search space. In our case of a 48-grid maze, the space of possible paths is therefore enormous. Generative models can struggle to maintain spatial consistency in such a large context window, often hallucinating \cite{wei2025satbench}, where example visualizations for our problem are shown in Fig. \ref{fig:vlm}. However, standard planning approaches (Trajectory Diffusion) often treat obstacles as soft constraints, which results in additional guidance procedures such as guidance or Langran methods \cite{liang2025simultaneous}. The main drawbacks of existing methods include the inaccuracy of gradient-based solutions and their high computational cost \cite{liang2025simultaneous} or difficulties in inferring spatial and logical relations using Multimodal Large Language Models (MLLM/VLM) \cite{makinski2025reasoning}.

This work proposes solutions to both of these problems by formulating path planning as a Conditional Image Generation task. Instead of relying on coordinate sequence optimization or the uncertain reasoning of VLMs, we propose a solution to the problem by operating solely on a four-channel image representation encoding information about the start, finish, and walls (obstacles). Our goal is to train a generative model that implicitly learns to avoid obstacles by considering their positions directly in the model's input layer.

\section{GenPlanner}
\begin{figure}[!t]
    \centering
    \includegraphics[width=1.0\linewidth]{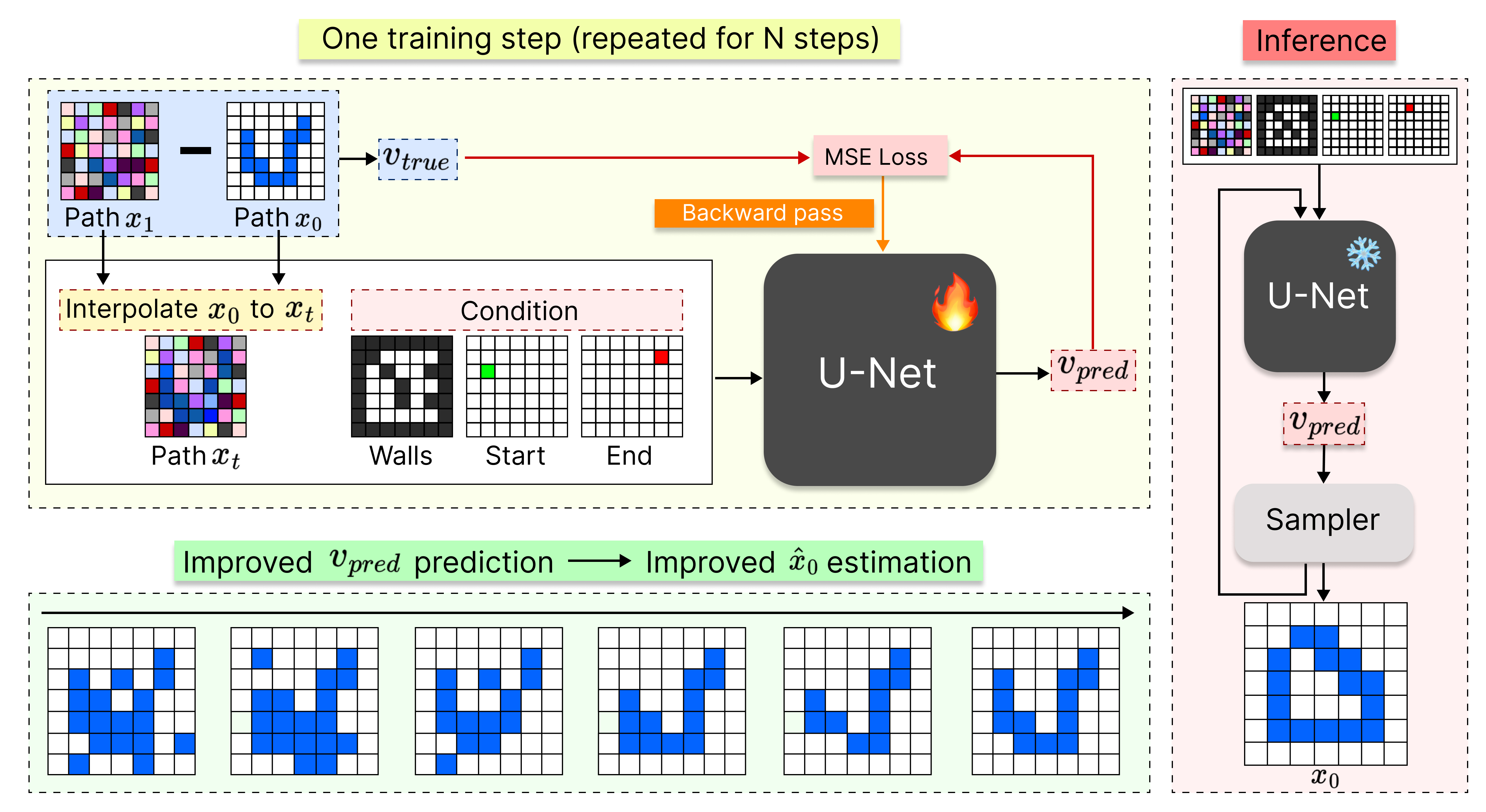}
    \caption{\textbf{Overview of FlowPlanner training and inference.} FlowPlanner treats path planning as a denoising process. A U-Net conditioned on walls and start-goal locations predicts velocity at each timestep, enabling iterative refinement from random noise to correct the route, where the pixel value > 0 is a path. DiffPlanner has a similar scheme where instead of operating on the velocity vector, the U-Net predicts noise.}
    \label{fig:method}
\end{figure}

This section introduces our \our{} method, a generative approach to planning based on conditional models. It includes two variants: DiffPlanner, which utilizes a diffusion model, and FlowPlanner, which is based on flow matching. \our{} is designed to construct paths in mazes and directly generates solutions in the form of consistent trajectories. An overview of the FlowPlanner is presented in Fig. \ref{fig:method}. Unlike popular text-to-image models (\cite{flux2024}, \cite{sdxl}), our method conditions on a multi-channel representation of the maze rather than on a textual description. First, we describe the input data representation, followed by an overview of the architecture of the proposed models. Finally, this section concludes with a full explanation of the training and testing process.
\subsection{Architecture}

\textbf{Input Data Representation}
The input to the generative model comprises two components: a noisy path map and a condition that describes the maze structure. The binary path map ${x}_0 \in \{0,1\}^{H \times W}$ is initially mapped to the range of $[-1,1]$ and then noisy by linear interpolation with a random noise sample ${\epsilon} \sim \mathcal{N}(0,I)$.  
For the diffusion model, the noised sample is constructed for a timestep $t \in \{0,\dots,T-1\}$ as follows:
\begin{equation}
{x}_t = \sqrt{\alpha_t}{x}_0 + \sqrt{1-\alpha_t}{\epsilon}.  
\end{equation}
where $T=1000$ in all experiments.
For the flow model, a sample is created for a randomly selected $t$ in the range $[0,1]$:
\begin{equation}
{x}_t = (1 - t){x}_0 + tx_1
\end{equation}
where in the flow matching formulation, the noise variable is denoted by $x_1$ instead of $\epsilon$.
The condition ${c} \in \{0,1\}^{3 \times H \times W}$ is the concatenation of three binary masks: the walls map, the starting point mask, and the destination point mask. The final input to the network is obtained by concatenating $x_t$ and $c$ along the channel dimension, resulting on a four-channel input tensor. In the inference phase, pure noise $\epsilon$ is supplied instead of $x_t$, and the model generates a path map conditioned solely on the maze structure.

\subsection{Training Objective}
\textbf{DiffPlanner Training}
During training, a forward diffusion process is simulated, in which Gaussian noise $\epsilon$ is gradually added to the binary path map $x_0$ for a specified timestep $t$, according to Eq. 1. The model, conditioned on the maze representation $c$, learns to predict noise based on the triplet ($x_t$, $c$, $t$), i.e., it estimates the noise ${\epsilon}_\theta({x}_t, {c}, t)$. Training is accomplished by minimizing the mean square error:
\begin{equation}
\mathcal{L}_{\mathrm{DiffPlanner}} = \mathbb{E}_{{x}_0, t, {\epsilon}}
\left[
\left\|
{\epsilon} 
- 
{\epsilon}_\theta({x}_t, {c}, t)
\right\|_2^2
\right].
\end{equation}
As a result, the model learns to effectively predict and remove noise while reconstructing a path that is consistent with the given maze.

\textbf{FlowPlanner Training}
FlowPlanner is based on learning continuous dynamics, which describes the evolution of a sample from random noise to the correct path. Instead of iteratively adding and removing noise, as in diffusion, a flow matching approach is used, in which the model directly learns the vector field that controls this transformation.
For each sample, a noise map ${x}_1 \sim \mathcal{N}(0,I)$ is sampled, and the ground-truth path mask $x_0$ is used. For a randomly sampled scalar $t$, an intermediate point is computed according to Eq.2.

This design method ensures that the model observes states with varying degrees of noise, from nearly clean paths to samples approaching random noise \cite{flow_train}. The learning goal is to approximate the velocity at which a sample should be moved towards the solution. The target vector field is defined as the time derivative of the interpolation trajectory: 
\begin{equation}
\frac{d{x}_t}{dt} ={x}_1 - {x}_0 
\end{equation}
The model, conditioned on the maze representation $c$, estimates the function ${v}_\theta({x}_t,{c}, t)$, which describes the local direction and rate of this transformation. The network parameters are optimized by minimizing the mean square error between the predicted and target velocities.
\begin{equation}
\mathcal{L}_{{\mathrm{FlowPlanner}}} =
\mathbb{E}_{{x}_0, {x}_1, t}
\left[
\left\|
(x_1-x_0)
-
{v}_\theta({x}_t,{c}, t)
\right\|_2^2
\right]
\end{equation}
As a result, FlowPlanner learns continuous dynamics that enable the gradual transformation of the random initialization into a correct and consistent path consistent with the maze structure.

\subsection{Inference}
\begin{figure}[t]
\centering
\setlength{\tabcolsep}{1pt}
\renewcommand{\arraystretch}{0.8}
\begin{tabular}{cccccc}
 & $\hat{{x}}_{0,50}$ &  $\hat{{x}}_{0,40}$ &  $\hat{{x}}_{0,30}$ &  $\hat{{x}}_{0,20}$ &  $\hat{{x}}_{0,0}$ \\
\rotatebox{90}{\scriptsize\hspace{6pt}{DiffPlanner}} &
\includegraphics[width=0.19\textwidth]{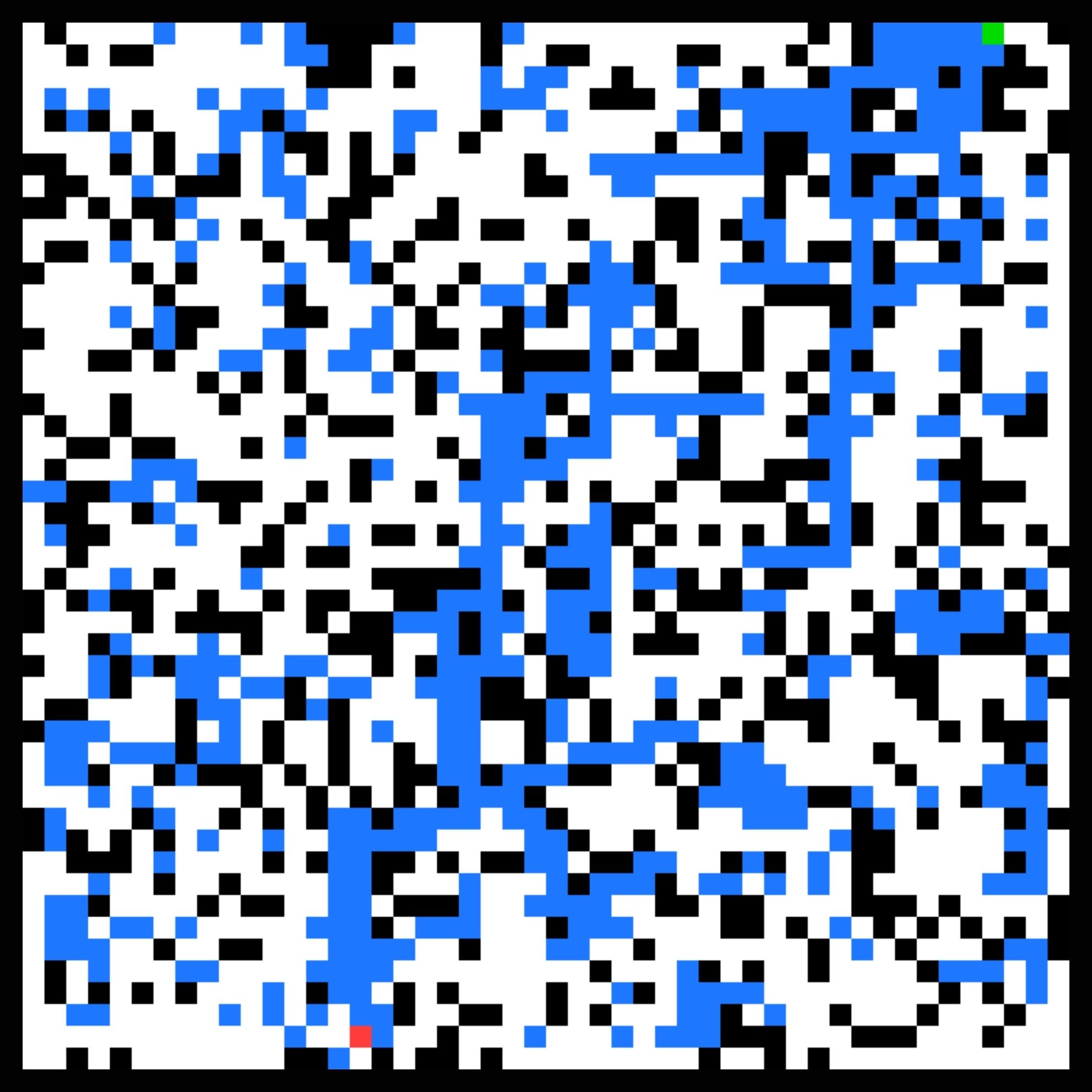} &
\includegraphics[width=0.19\textwidth]{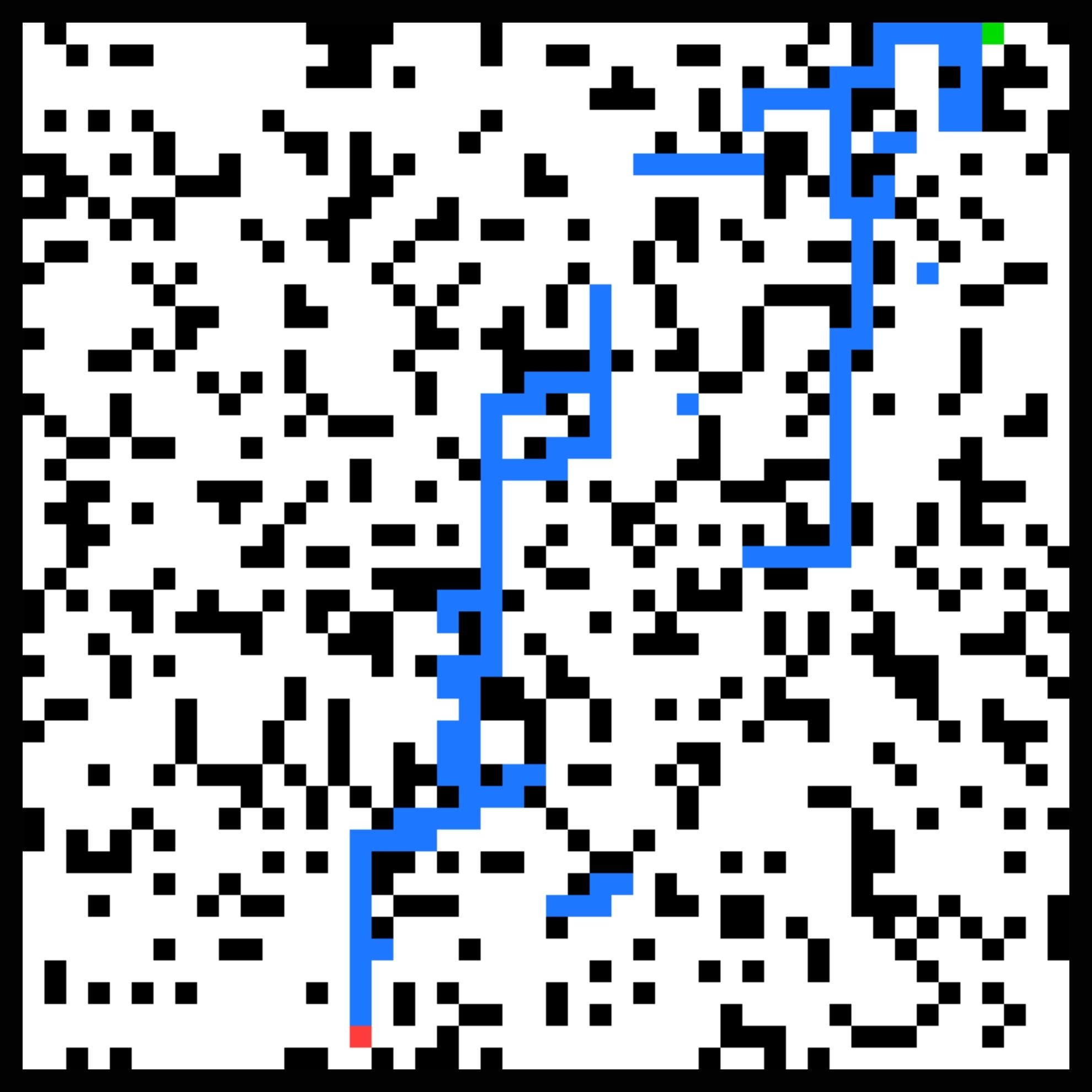} &
\includegraphics[width=0.19\textwidth]{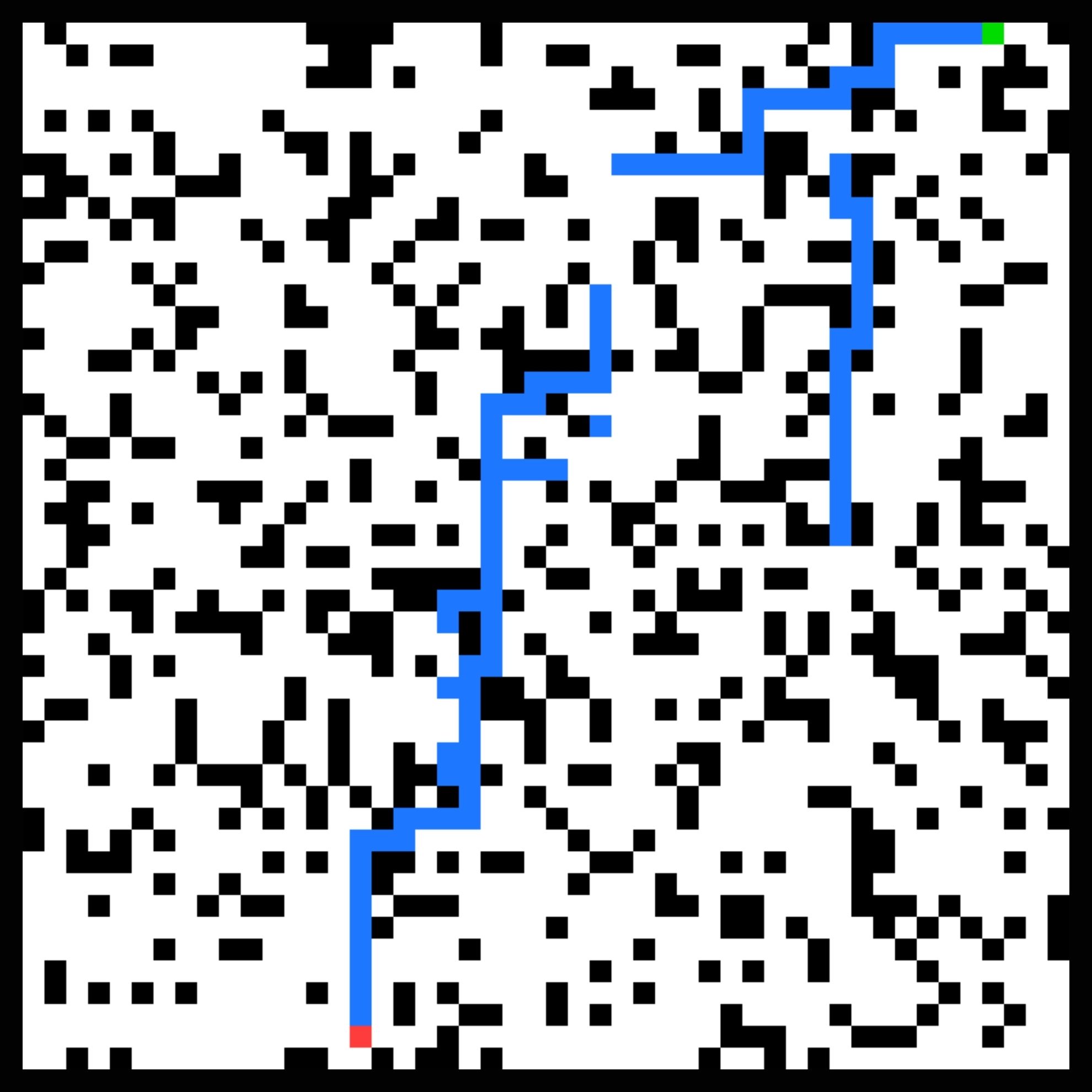} &
\includegraphics[width=0.19\textwidth]{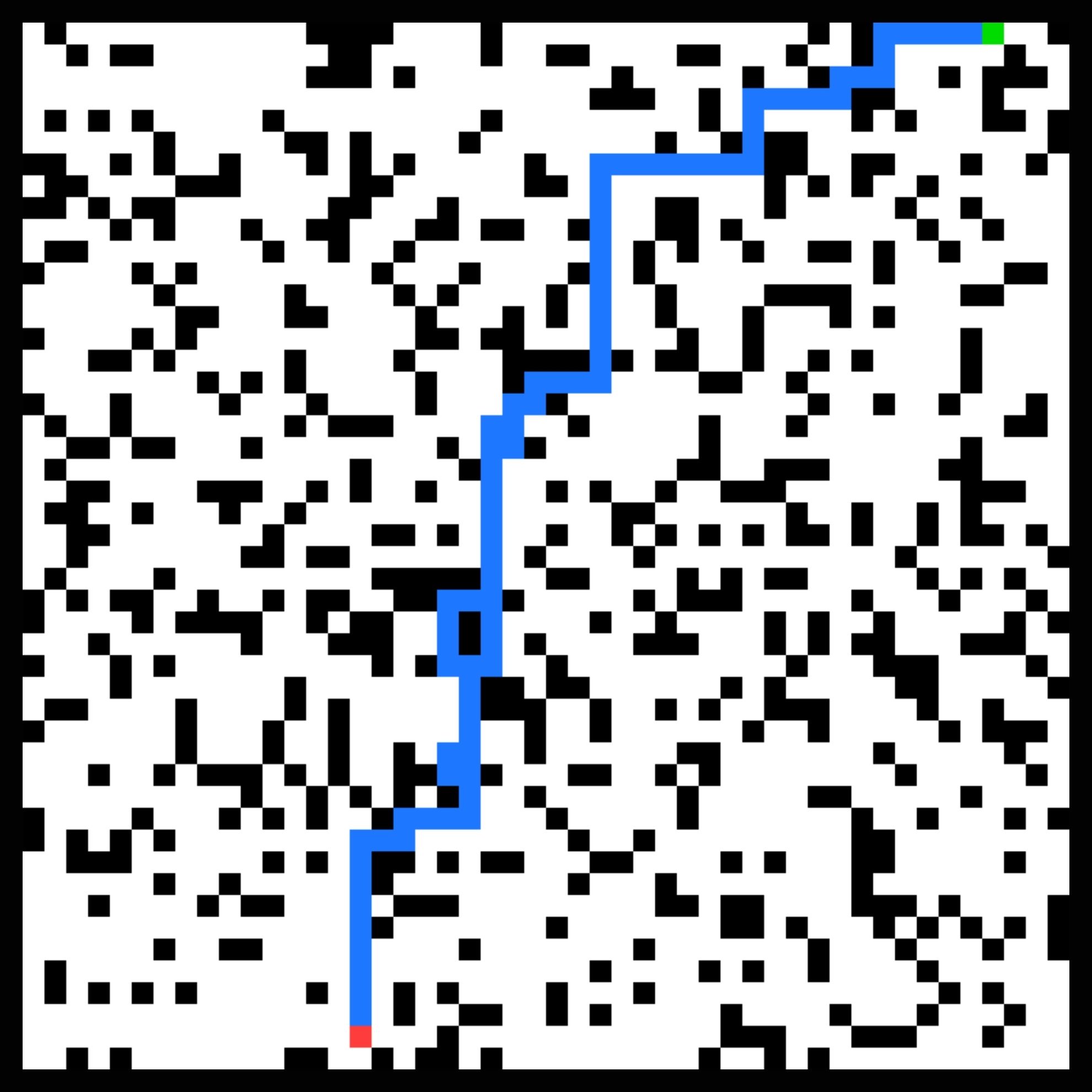} &
\includegraphics[width=0.19\textwidth]{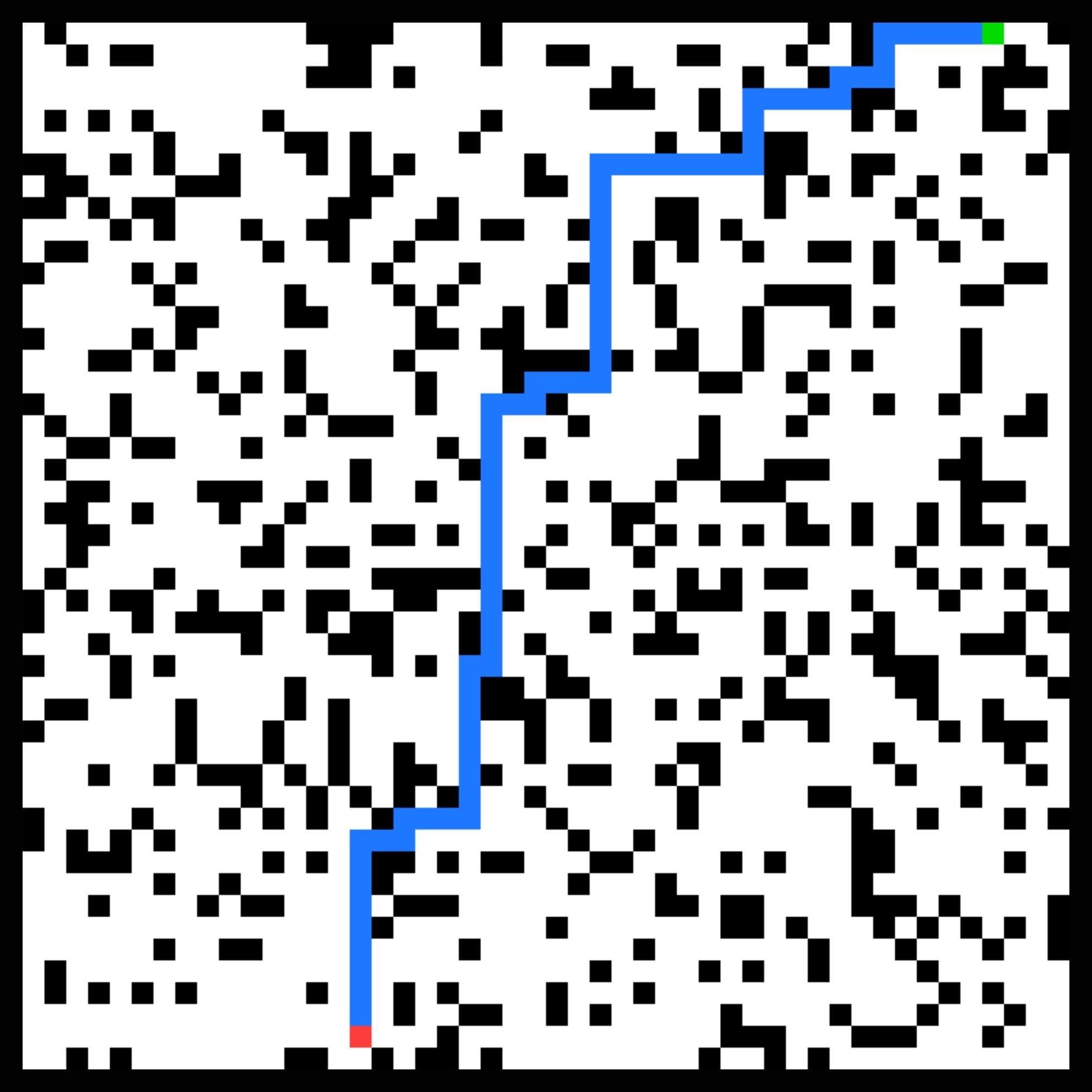} \\
\rotatebox{90}{\scriptsize\hspace{4pt}{FlowPlanner}} &
\includegraphics[width=0.19\textwidth]{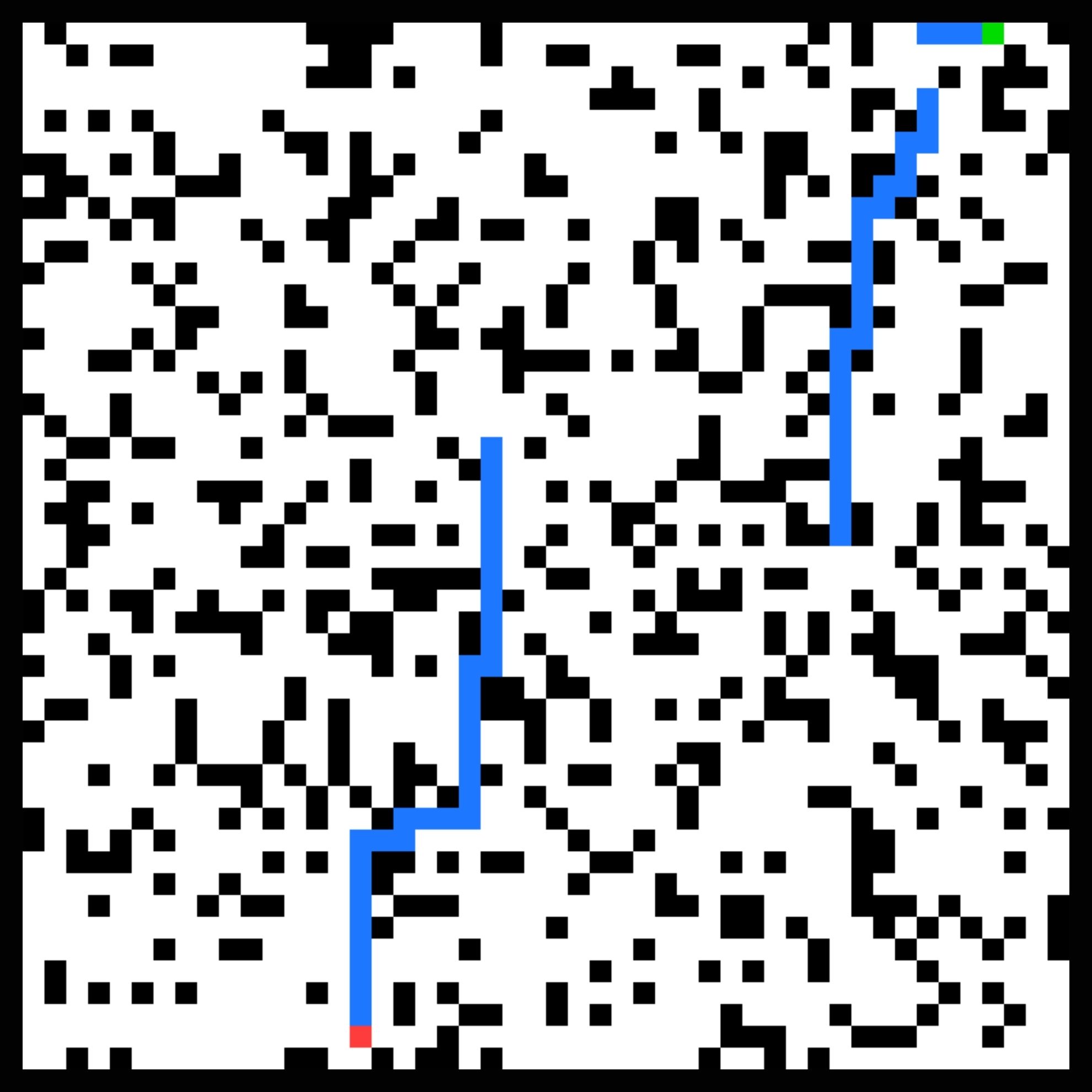} &
\includegraphics[width=0.19\textwidth]{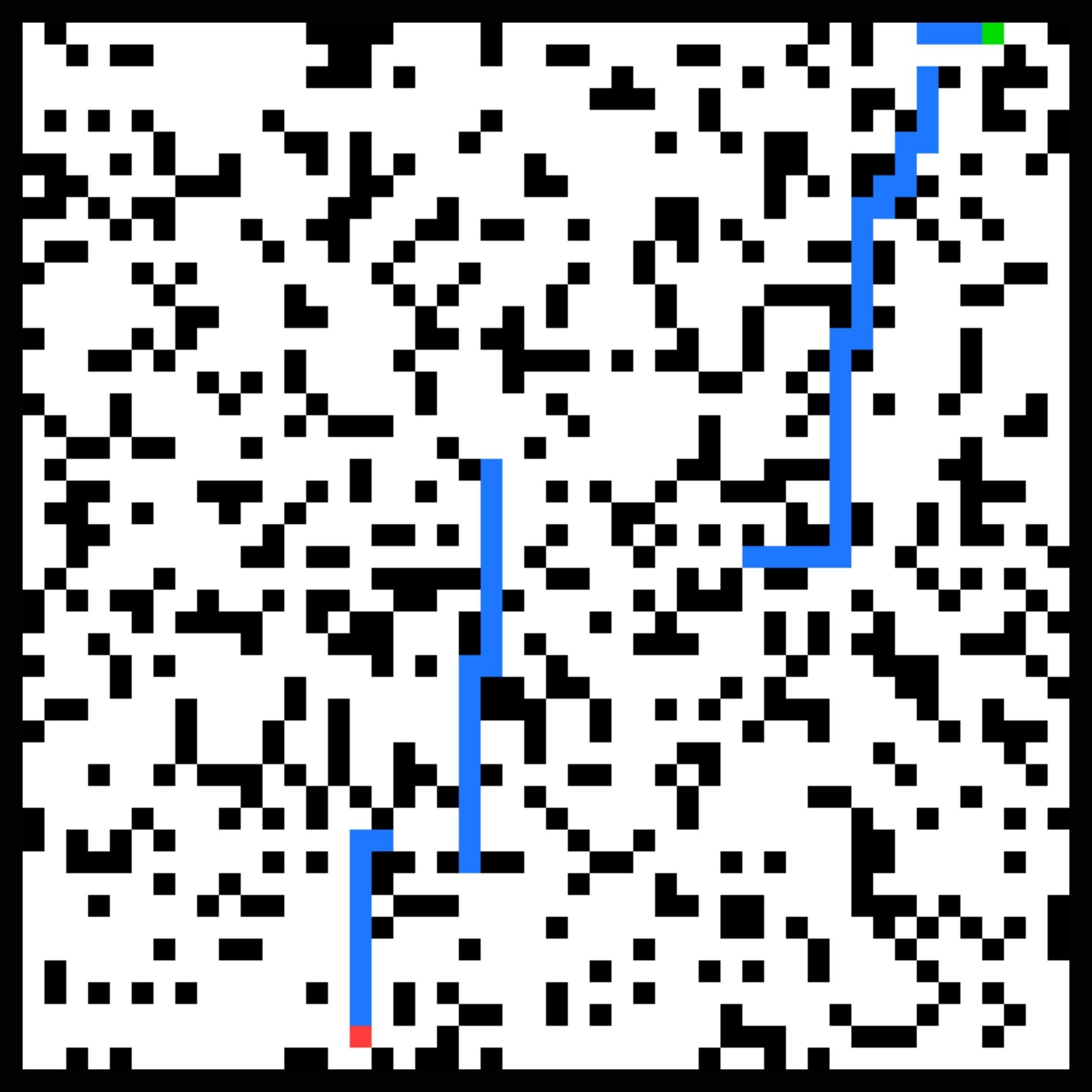} &
\includegraphics[width=0.19\textwidth]{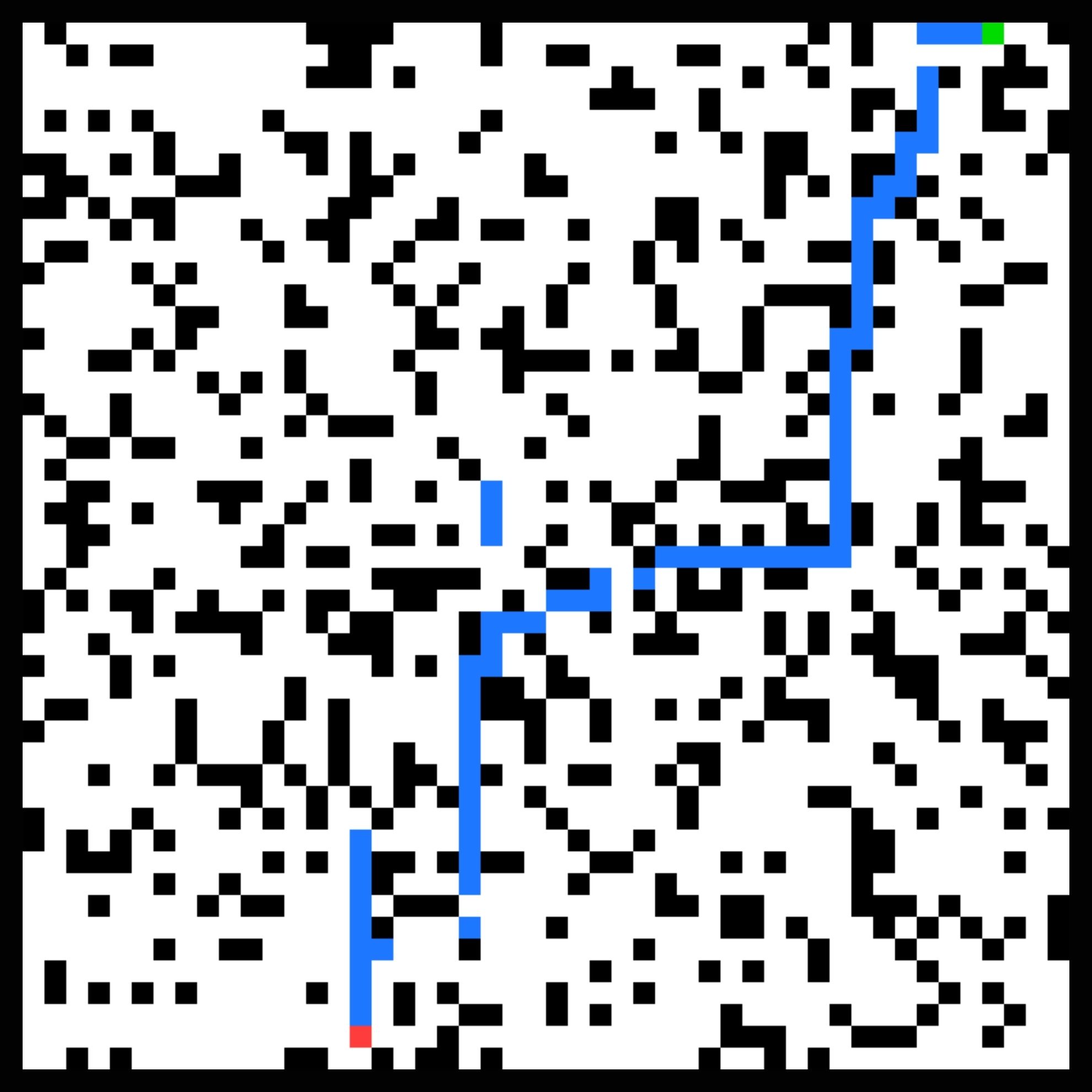} &
\includegraphics[width=0.19\textwidth]{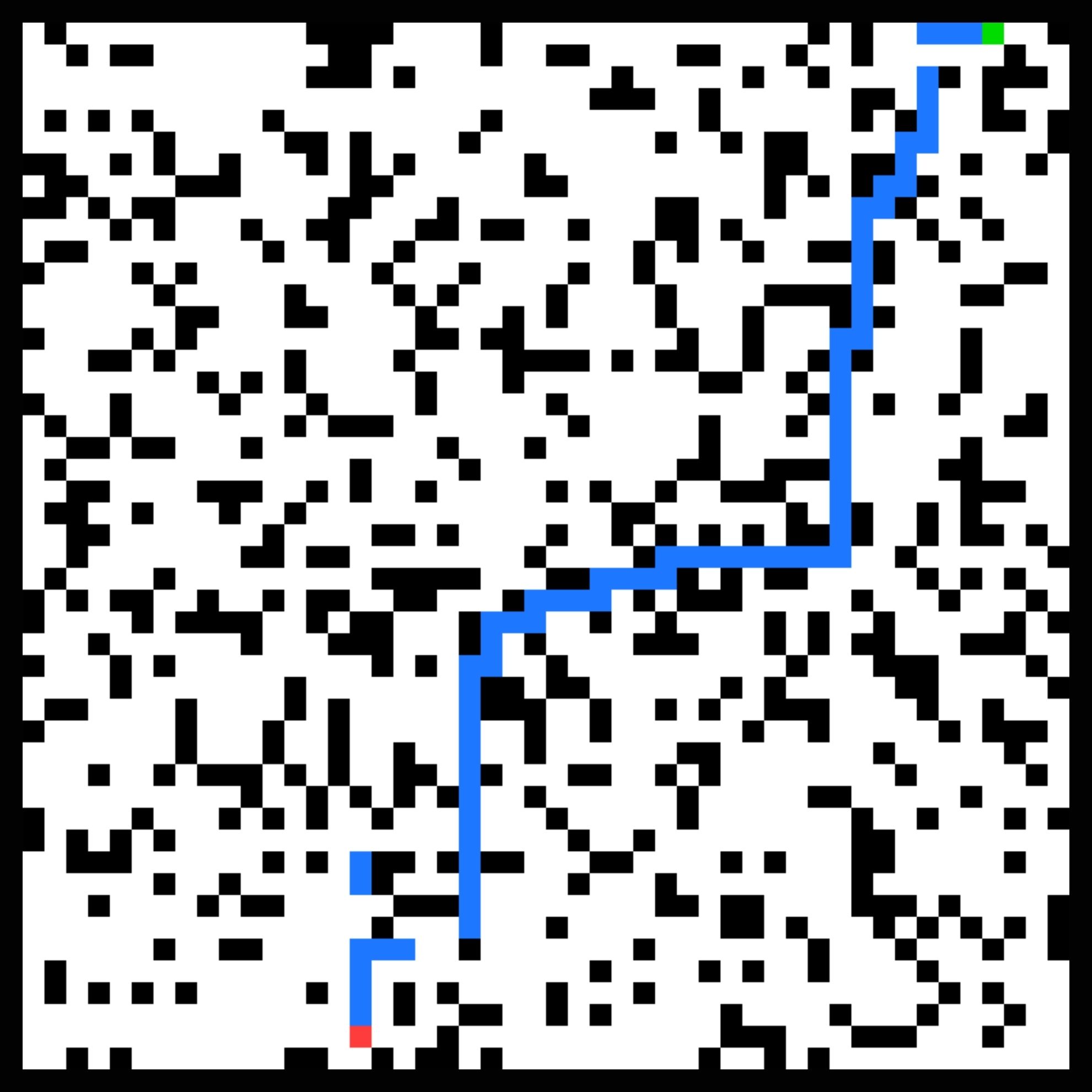} &
\includegraphics[width=0.19\textwidth]{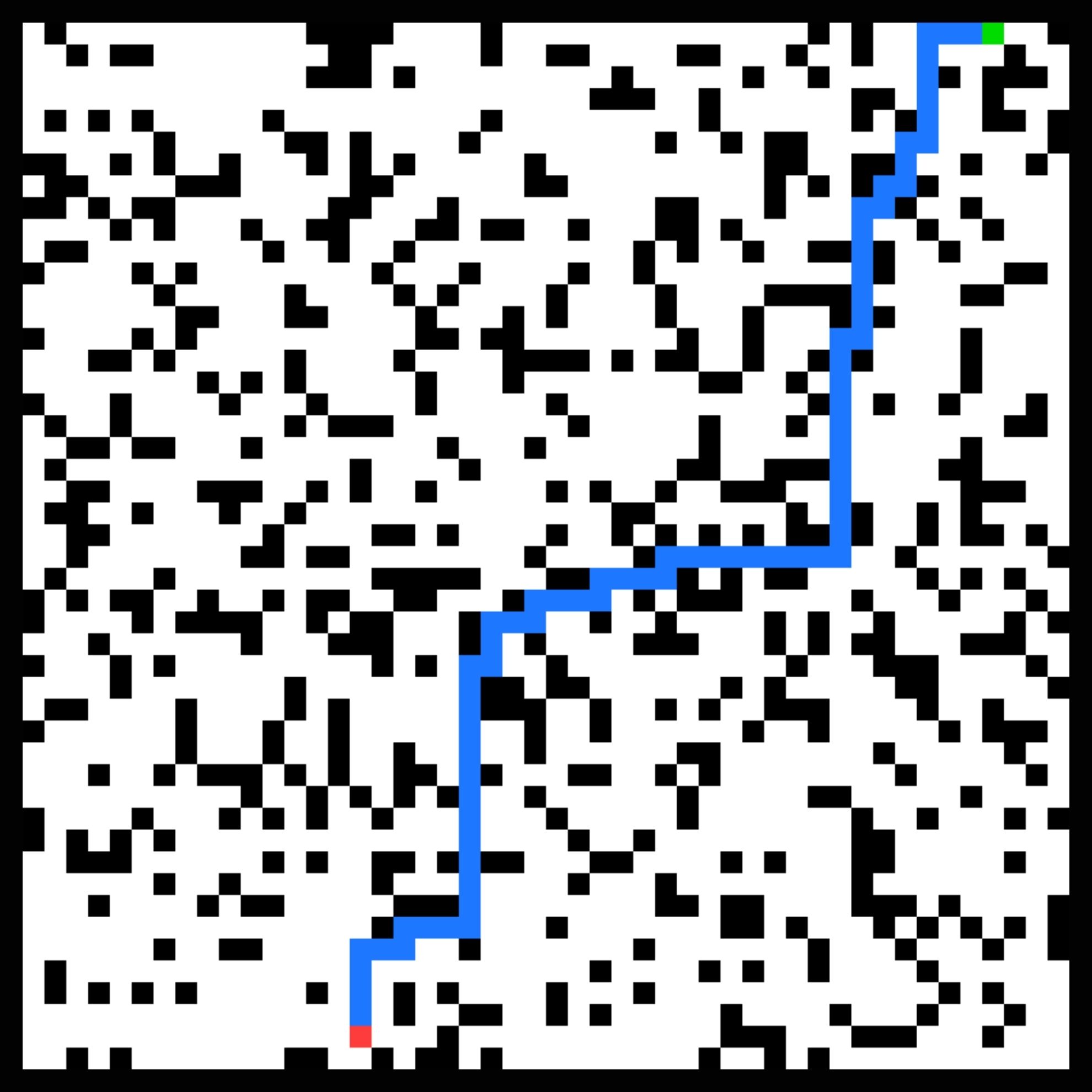} \\
\end{tabular}
\caption{\textbf{Intermediate $\hat{{x}}_{0,t}$ Estimates for Diffusion (top) and Flow Models (bottom).} For DiffPlanner, the path structure is formed gradually with subsequent denoising steps, while FlowPlanner produces coherent trajectory structures already at early integration steps, as the predicted velocity field directly guides the sample toward the final solution.}
\label{fig: x0}
\end{figure}
\textbf{FlowPlanner inference} Path generation is formulated as a solution to an ordinary differential equation describing continuous dynamics transforming random noise into a solution. In the inference phase, FlowPlanner starts generation with a random noise sample ${x}_1 \sim \mathcal{N}(0,I)$. Then, the model, conditional on the maze structure ${c}$, estimates the vector field ${v}_\theta({x}_t, {c}, t)$ describing the instantaneous evolution rate of the sample. Based on this velocity, the Sampler (see Fig. \ref{fig:method}) updates the state using the explicit Euler schema:
\begin{equation}
 {x}_{t+\Delta t}
=
{x}_t
+
\Delta t\,{v}_\theta({x}_t, {c}, t)   
\end{equation}
where $\Delta t <0$ corresponds to backward integration in time from $t=1$ to $t=0$, with the interval $[0,1]$ discretized into $T$ uniform integration steps. This process is repeated iteratively until a final state is obtained corresponding to the path map.

As in diffusion models, the flow-based approach also allows for the explicit estimation of a pure sample at any time step. For each $x_t$ during generation and the predicted velocity field ${v}_\theta({x}_t,{c},t)$ the solution estimation can be written as:
\begin{equation}
\hat{{x}}_{0,t}
= {x}_t
-
t\,{v}_\theta({x}_t,{c},t)
\end{equation}
Unlike diffusion, where $\hat{{x}}_{0,t}$ is recovered by inverting the noise process, FlowPlanner calculates the solution directly based on the current state and predicted velocity, leading to more stable intermediate estimates (see Fig. \ref{fig: x0}).

\textbf{DiffPlanner inference}
Path generation is implemented via an inverse diffusion process, in which random noise is gradually transformed into a solution map. The process begins with a Gaussian random noise sample ${x}_T \sim \mathcal{N}(0,I)$. For the subsequent time steps $t = T-1 \dots, 0$, the model, conditioned on the maze structure $c$, predicts the additional noise ${\epsilon}_\theta({x}_t, {c}, t)$ and then updates the state. During inference, a deterministic DDIM-sampler \cite{ddim_sampler} is used instead of the stochastic DDPM sampler \cite{ddpm}. Unlike DDPM, which adds additional random noise to the sample at each step, the DDIM sampler does not introduce new random perturbations but uses only the predicted noise to determine the next state. This allows for fewer generation steps and flexible scaling of the inference process without compromising the quality of the results. First, the clean sample is estimated as:
\begin{equation}
\hat{{x}}_{0,t} =
\frac{{x}_t -
\sqrt{1-\alpha_t}\,
{\epsilon}_\theta({x}_t, {c}, t)}
{\sqrt{\alpha_t}}.
\end{equation}
Next, the transition to the subsequent timestep is performed according to:
\begin{equation}
{x}_{t-1}
=
\sqrt{\alpha_{t-1}}\,
\hat{{x}}_{0,t}
+
\sqrt{1-\alpha_{t-1}}\,
{\epsilon}_\theta({x}_t, {c}, t).
\end{equation}
In our work, the DDIM sampler was adopted due to the possibility of selecting a smaller number of steps in the inference process. The effect of choosing the number of steps is shown in the Table. \ref{hyperparaemtrs_steps}. Notably, during generation, an estimate of the clean solution can be computed at any timestep, as shown in Eq.7.  This estimate offers an intermediate prediction of the final path, indicating that the model refines its solution progressively as $t$ decreases. (see Fig. \ref{fig: x0}).

\section{Experiments}
\begin{table}[!t]
\centering
\caption{\textbf{Statistics of the generated datasets for different grid sizes.}  }
\setlength{\tabcolsep}{12pt}
{\fontsize{9.5pt}{11pt}\selectfont
\begin{tabular}{c|c|c|c}
\hline
Grid size      & Training samples & Evaluation samples & Min. path length \\ \hline
$48 \times 48$ & 20,000           & 1,000              & 20               \\
$32 \times 32$ & 20,000           & 1,000              & 10               \\
$16 \times 16$ & 10,000           & 500                & 5                \\
$8 \times 8$   & 5,000            & 250                & 1                \\ \hline
\end{tabular}
}
\label{tab:datasets}
\end{table}
In this section, we present both visual and quantitative experimental results and compare them with those obtained by other methods. We also describe the dataset creation process and discuss the metrics used for evaluation in detail. Finally, we present the results of ablation studies, analyzing the impact of individual model components on its performance.

\subsection{Dataset}
The dataset was generated by randomly creating mazes with specific sizes and obstacle densities. For each sample, a binary obstacle map is created in which each cell is randomly designated as a wall based on a specified probability. The starting and finishing points are then randomly selected from the available free grid cells. A shortest path algorithm, specifically A*, is used to find the solution that connects the starting point to the goal. Samples where no valid paths can be found are discarded, ensuring that each instance has a unique solution.

Four binary masks are saved for each accepted sample: the obstacle map, the starting point map, the goal point mask, and the solution path mask. The dataset is generated for various grid sizes ($8 \times8$, $16 \times16$, $32\times32$, $48\times48$), allowing for the analysis of model performance as sample sizes increase. Furthermore, for smaller meshes, fewer examples are generated because the space of possible configurations is significantly smaller. Detailed statistics of the datasets are presented in Table \ref{tab:datasets}.

\subsection{Metrics}
Four metrics were used to assess the quality of the generated paths. The correctness of the solution is measured by the \textit{Validity} metric, which checks whether the generated path forms a continuous connection between the starting point and the goal. The \textit{Single-Path} metric assesses whether the generated trajectory forms a single, unbranched path. It takes a value of 1 only if \textit{Validity}=1, the path contains no branches (\textit{Branch-Rate}=0), and has exactly two endpoints. The structural quality of the path is also assessed by the \textit{Branch-Rate}, which measures the percentage of the cells in the path with at least three neighbors and is reported as an additional diagnostic metric. Lower values of this metric indicate more regular trajectories. The efficiency of the generated solution is measured using the \textit{Length Ratio}, defined as the ratio of the length of the generated path to the length of the shortest path determined by the A* algorithm \cite{a*}.

\section{Results}
\begin{table}[!t]
 \centering
\caption{\textbf{Quantitative comparison of CNN, DiffPlanner, and FlowPlanner across different grid sizes.} The best results are achieved by FlowPlanner, which leverages a learned vector field to guide the sample along a continuous trajectory toward the solution.}
\setlength{\tabcolsep}{3pt}
{\fontsize{6.6pt}{6pt}\selectfont
\begin{tabular}{llcccc}
\toprule
Maze Size & Model & Validity (\%) $\uparrow$ & Single-Path (\%) $\uparrow$ & Length Ratio $\downarrow$ & Branch-Rate (\%) $\downarrow$ \\
\midrule

\multirow{3}{*}{8$\times$8}
 & CNN         & 92.80 & 89.20 & \textbf{1.00} & 0.29 \\
 & FlowPlanner & \textbf{94.00} & \textbf{92.40} & \textbf{1.00} & \textbf{0.26 }\\
 & DiffPlanner & 90.40 & 77.20 & 1.02 & 1.95 \\
\midrule

\multirow{3}{*}{16$\times$16}
 & CNN         & 74.00 & 65.20 & \textbf{1.00} & 0.82 \\
 & FlowPlanner & \textbf{88.60} & \textbf{86.20} & 1.01 & \textbf{0.19} \\
 & DiffPlanner & 84.60 & 67.60 & 1.03 & 2.15 \\
\midrule

\multirow{3}{*}{32$\times$32}
 & CNN         & 49.60 & 45.20 & \textbf{1.00} & 0.42 \\
 & FlowPlanner & 82.20 & 81.60 & 1.01 & \textbf{0.05} \\
 & DiffPlanner & \textbf{88.60} & \textbf{82.80} & 1.03 & 0.44 \\
\midrule

\multirow{3}{*}{48$\times$48}
 & CNN         & 38.70 & 28.40 & \textbf{1.00} & 0.95 \\
 & FlowPlanner & 88.00 & \textbf{86.10} & 1.02 & \textbf{0.09} \\
 & DiffPlanner & \textbf{89.00 }& 76.10 & 1.04 & 0.47 \\
\bottomrule
\end{tabular}
}
\label{tab:cnn_diff_flow}
\end{table}

\begin{figure*}[!t]
\centering
\setlength{\tabcolsep}{1.2pt}
\renewcommand{\arraystretch}{0.9}
\begin{tabular}{cccc}
Ground Truth &  CNN & DiffPlanner & FlowPlanner \\
\includegraphics[width=0.24\textwidth]{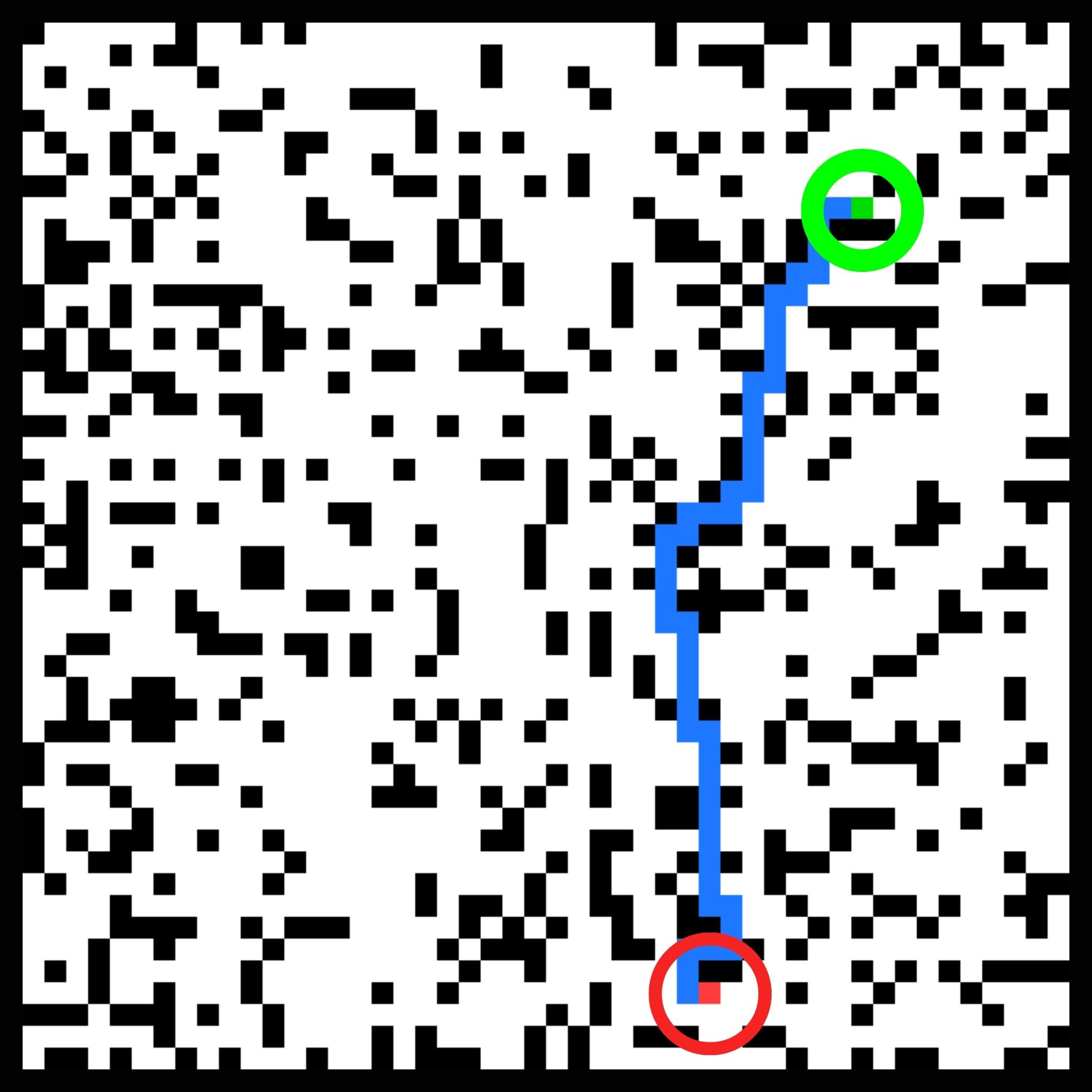} &
\includegraphics[width=0.24\textwidth]{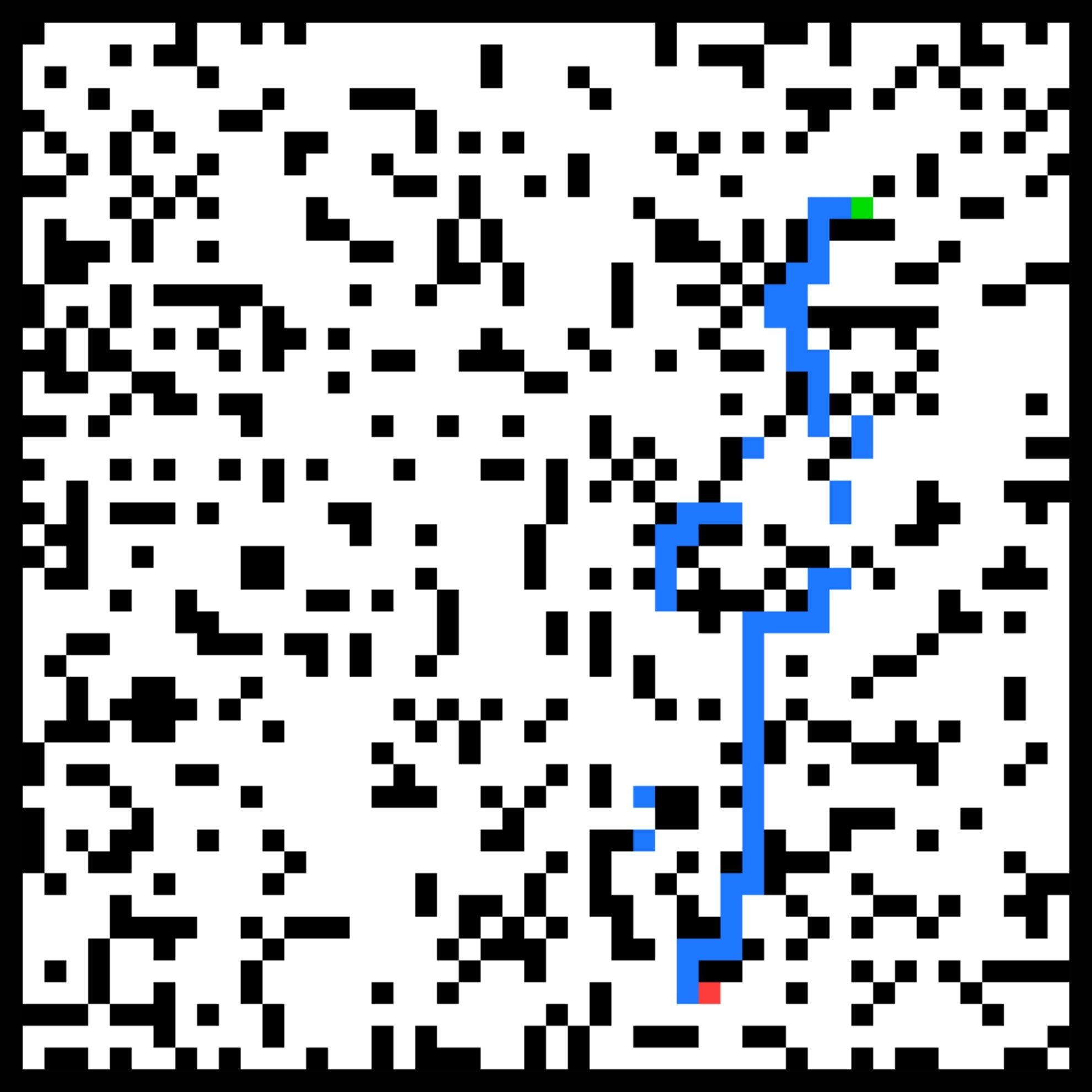} &
\includegraphics[width=0.24\textwidth]{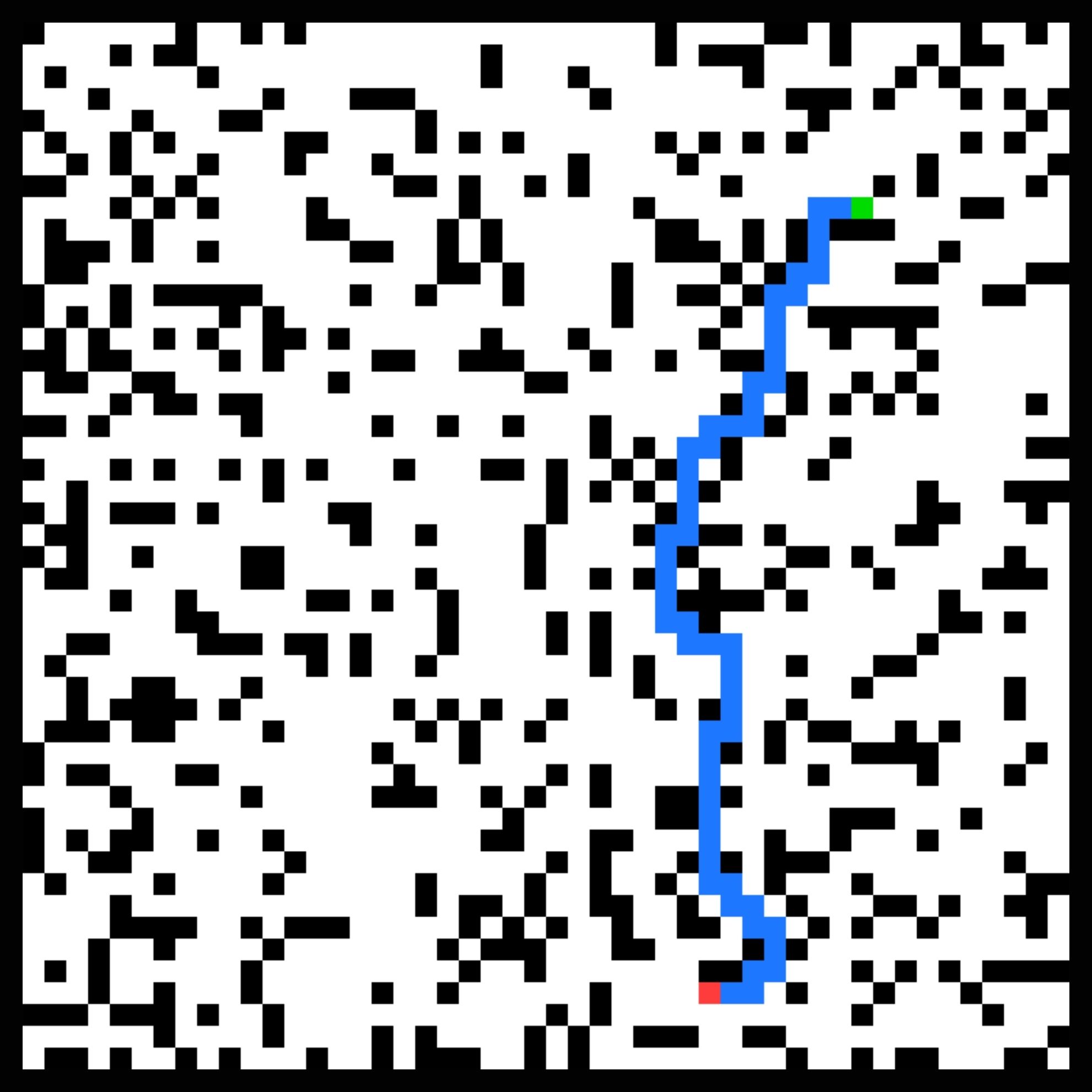} &
\includegraphics[width=0.24\textwidth]{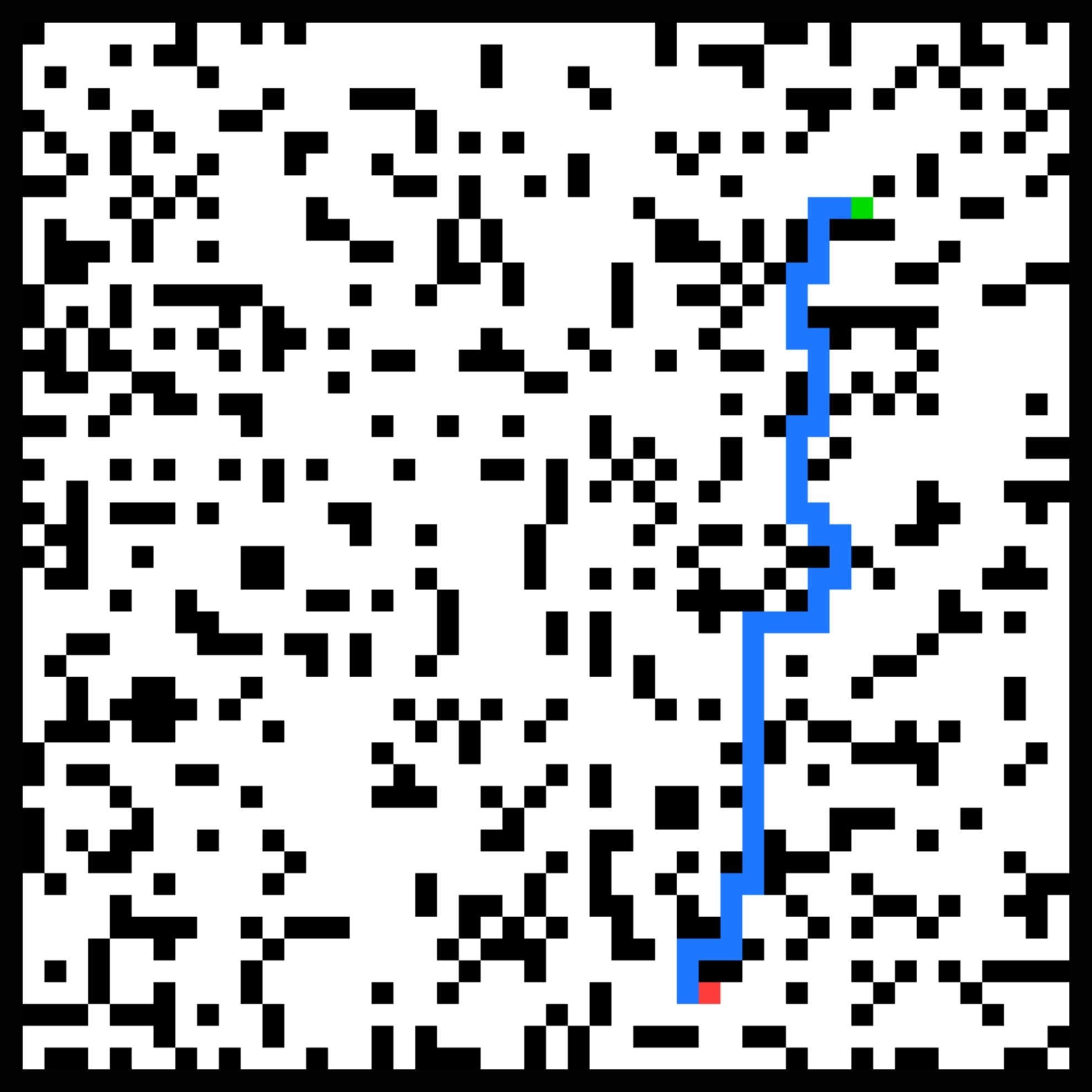} \\
\includegraphics[width=0.24\textwidth]{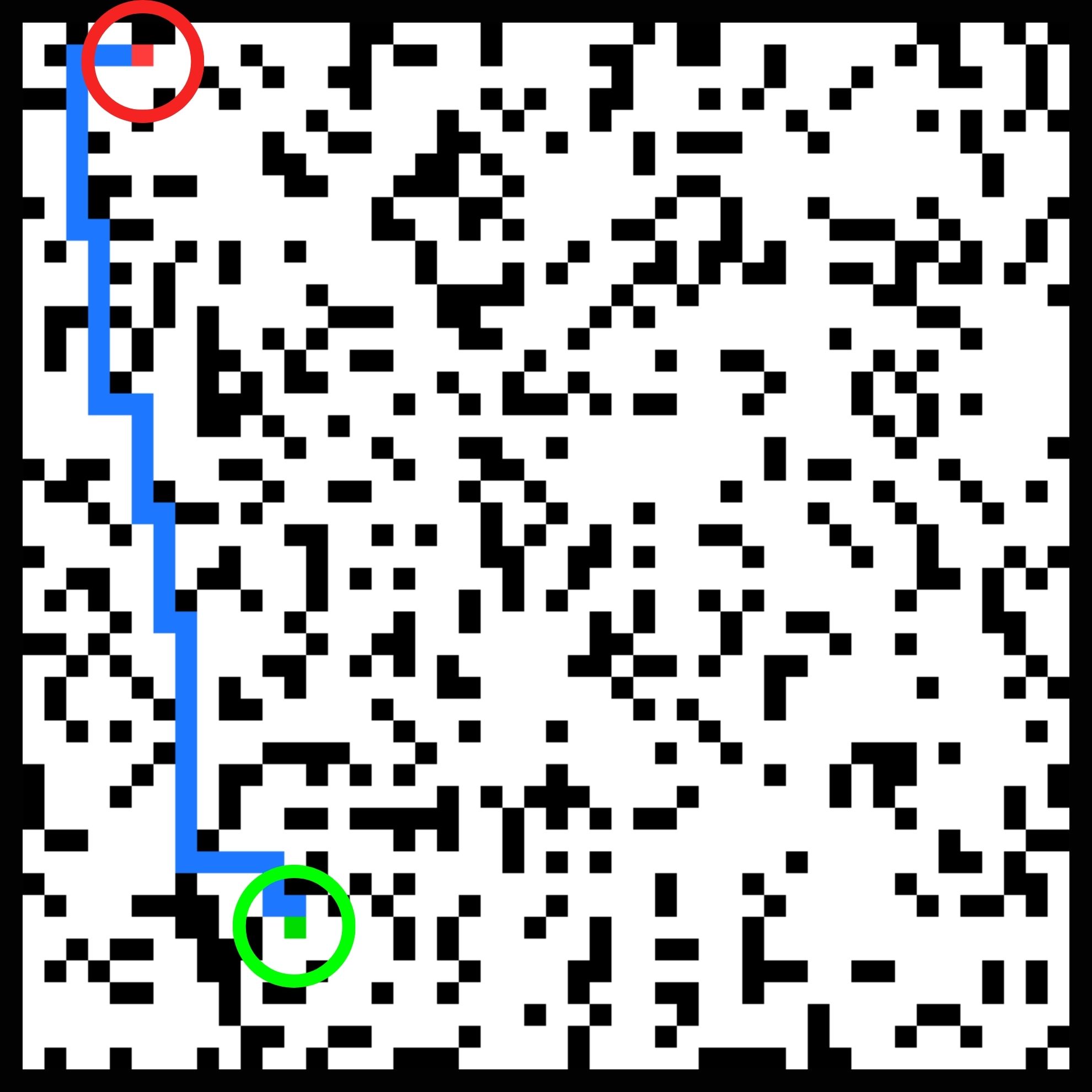} &
\includegraphics[width=0.24\textwidth]{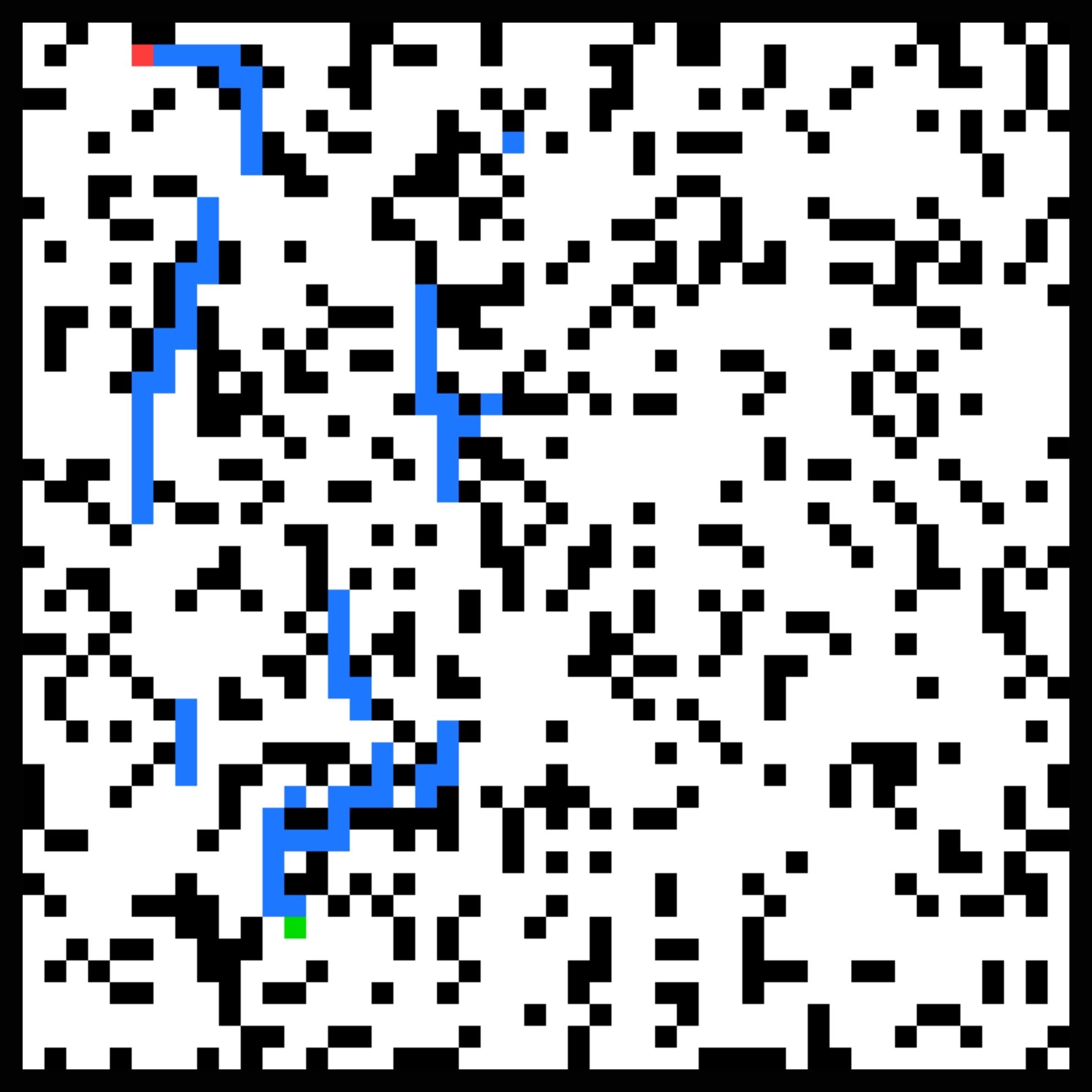}  &
\includegraphics[width=0.24\textwidth]{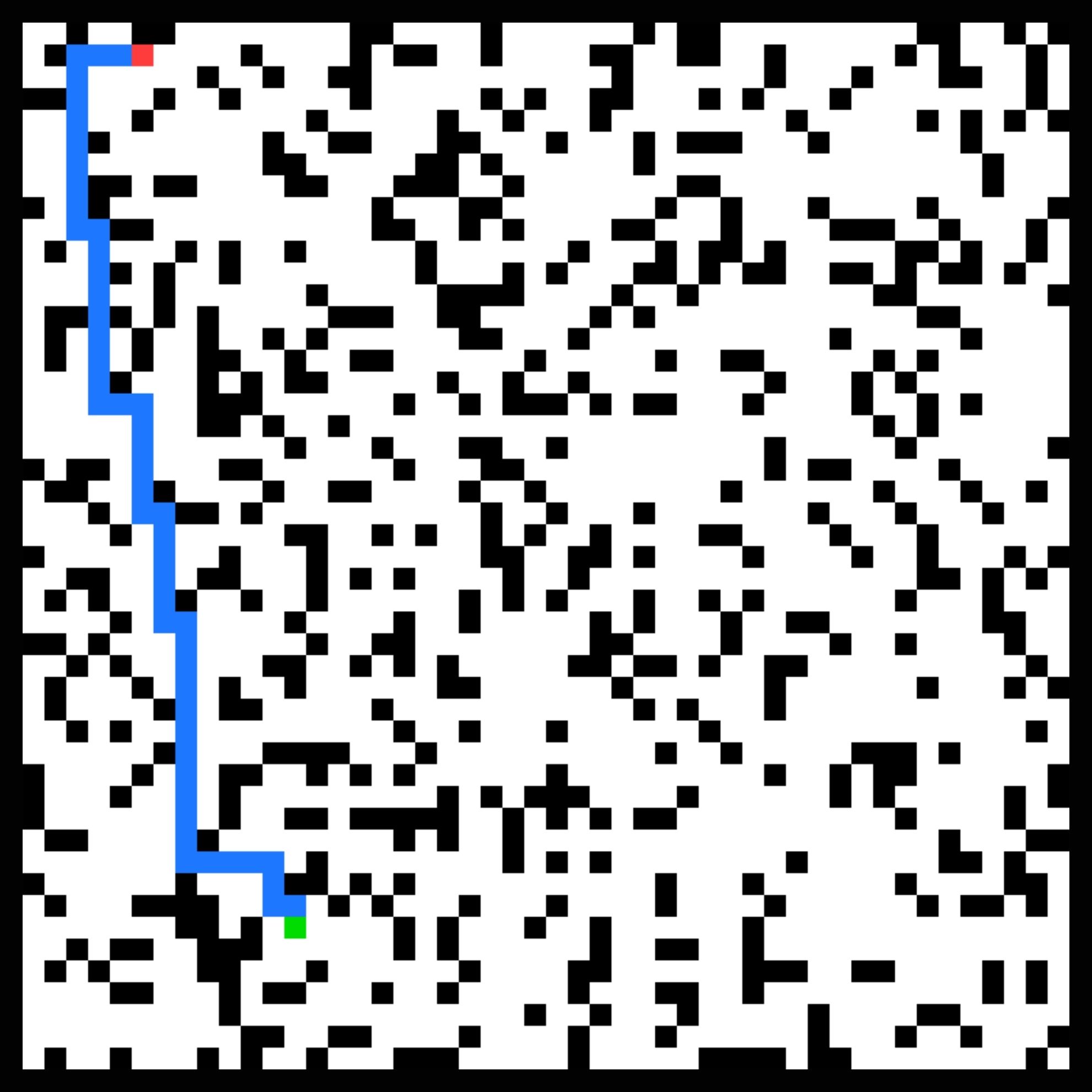} &
\includegraphics[width=0.24\textwidth]{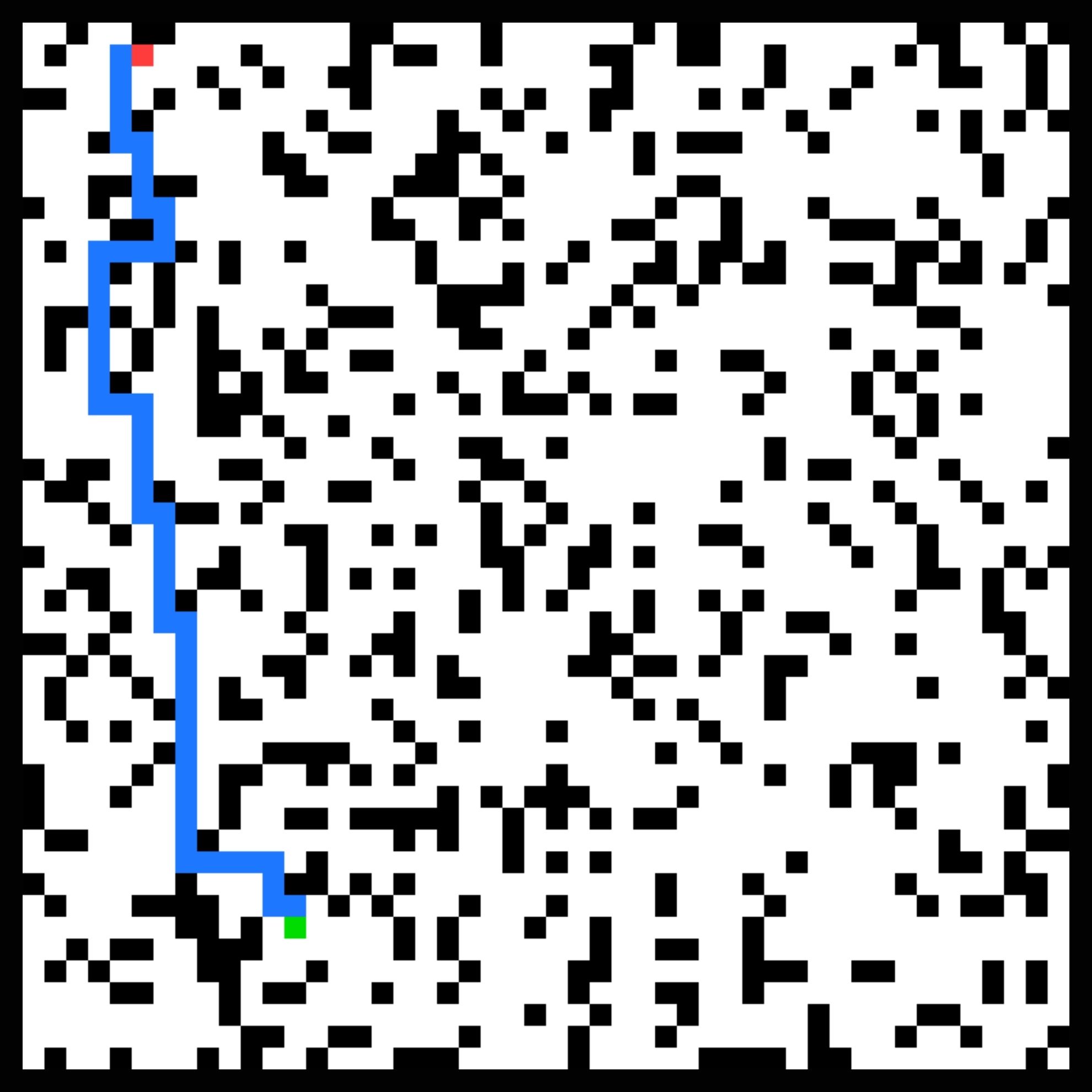}
\end{tabular}
\caption{\textbf{Qualitative comparison of paths generated by CNN, DiffPlanner, and FlowPlanner on $48\times48$ mazes.}}
\label{fig:48x48}
\end{figure*}

\begin{figure*}[!t]
\centering
\setlength{\tabcolsep}{1.2pt}
\renewcommand{\arraystretch}{0.9}
\begin{tabular}{cccc}
Ground Truth &  CNN & DiffPlanner & FlowPlanner \\
\includegraphics[width=0.24\textwidth]{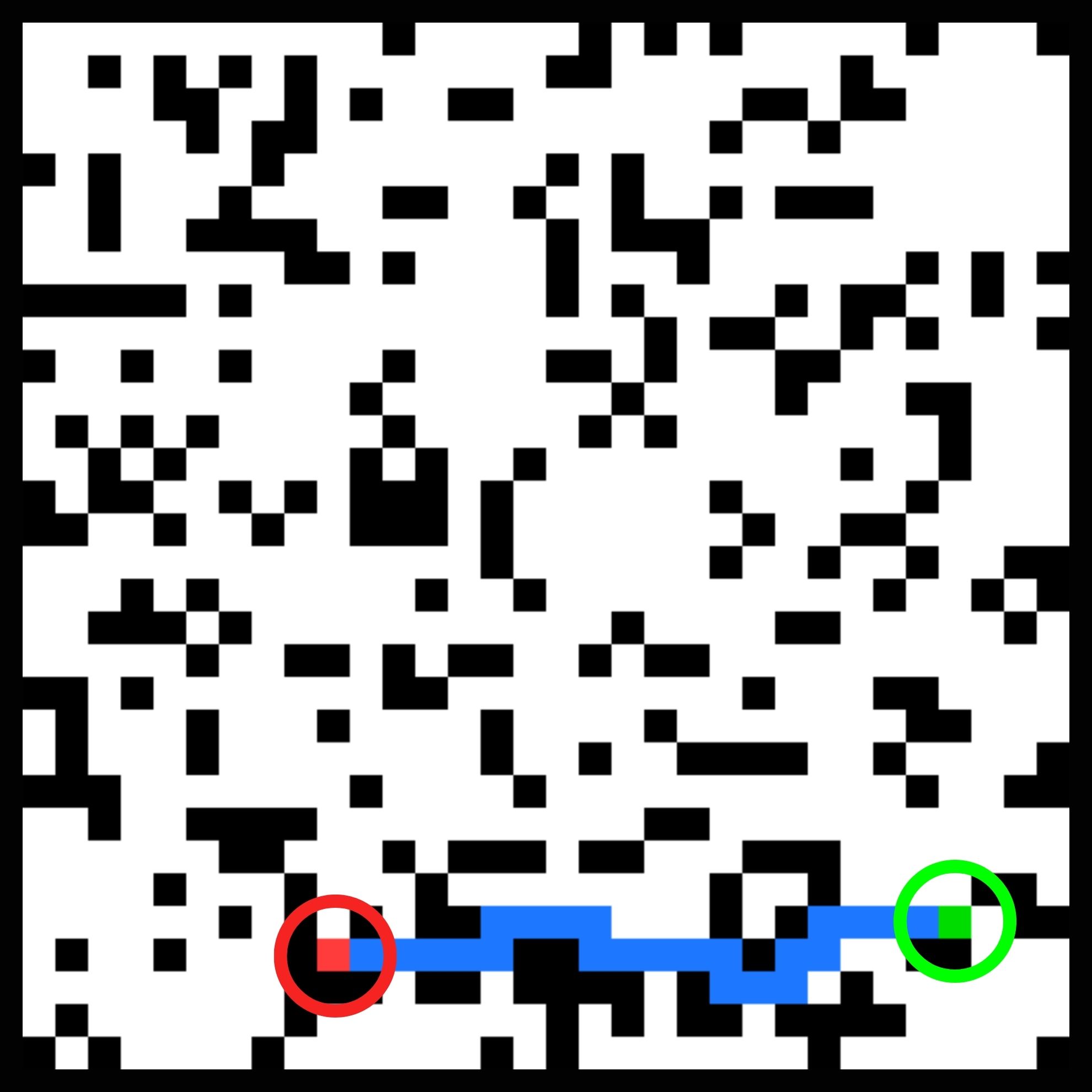} &
\includegraphics[width=0.24\textwidth]{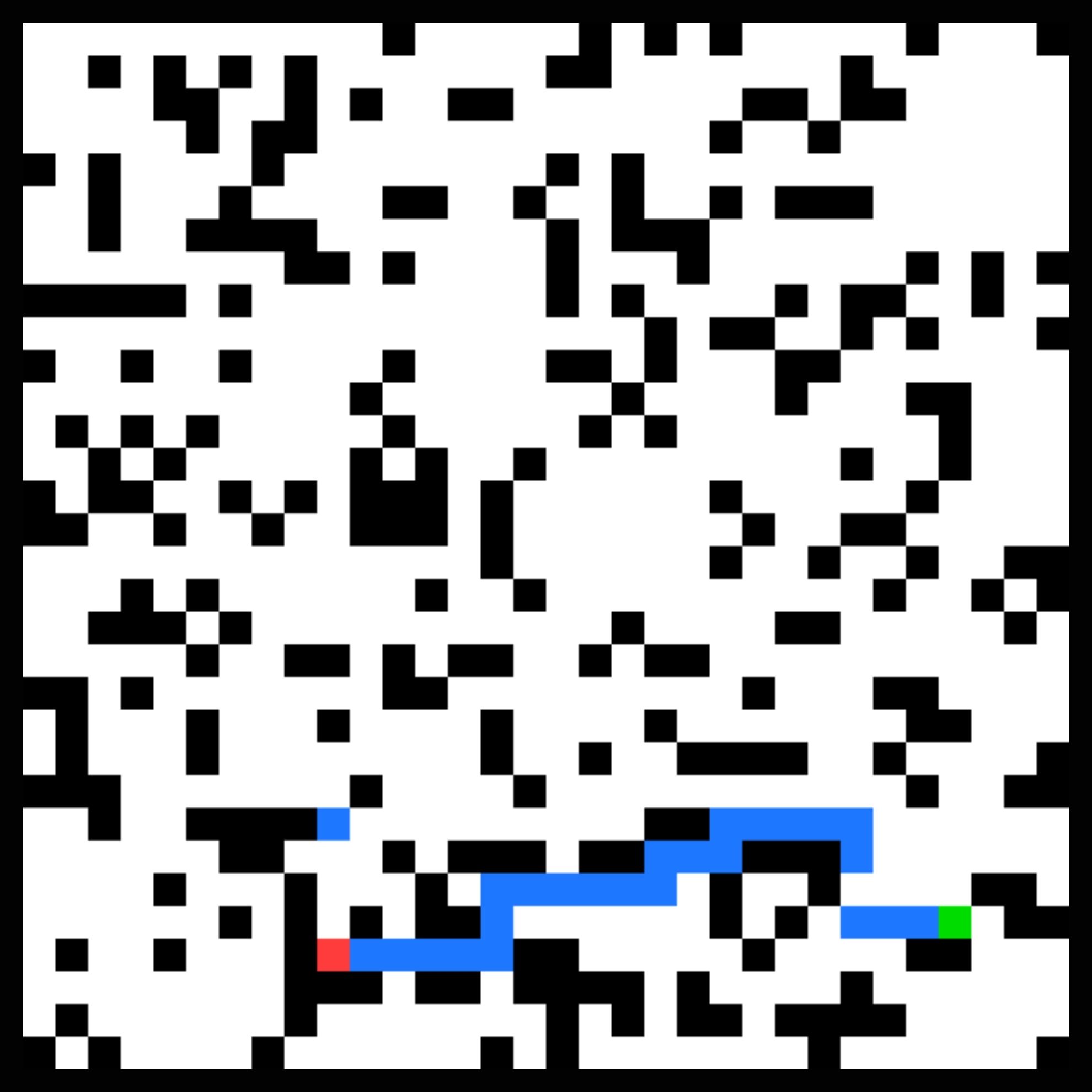} &
\includegraphics[width=0.24\textwidth]{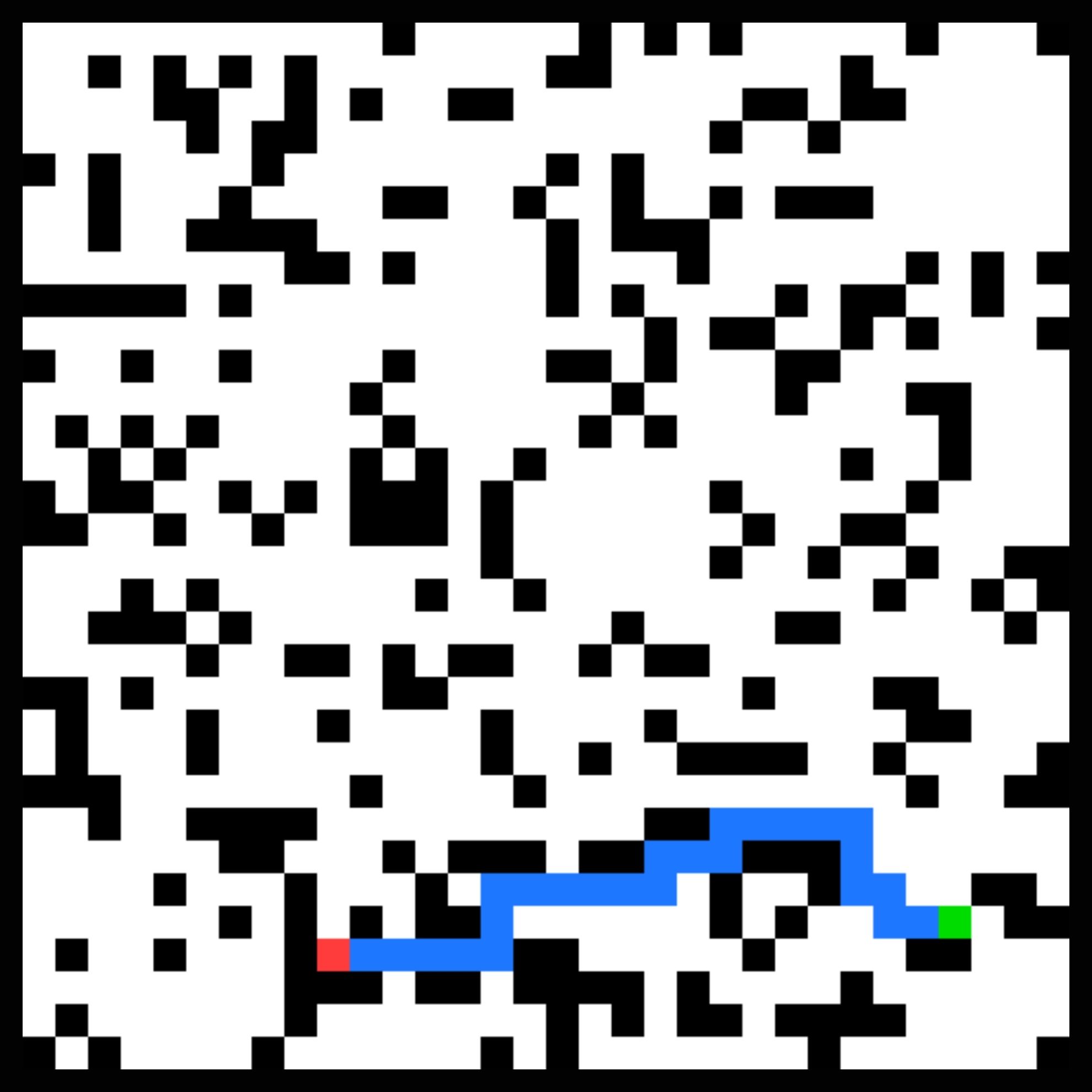} &
\includegraphics[width=0.24\textwidth]{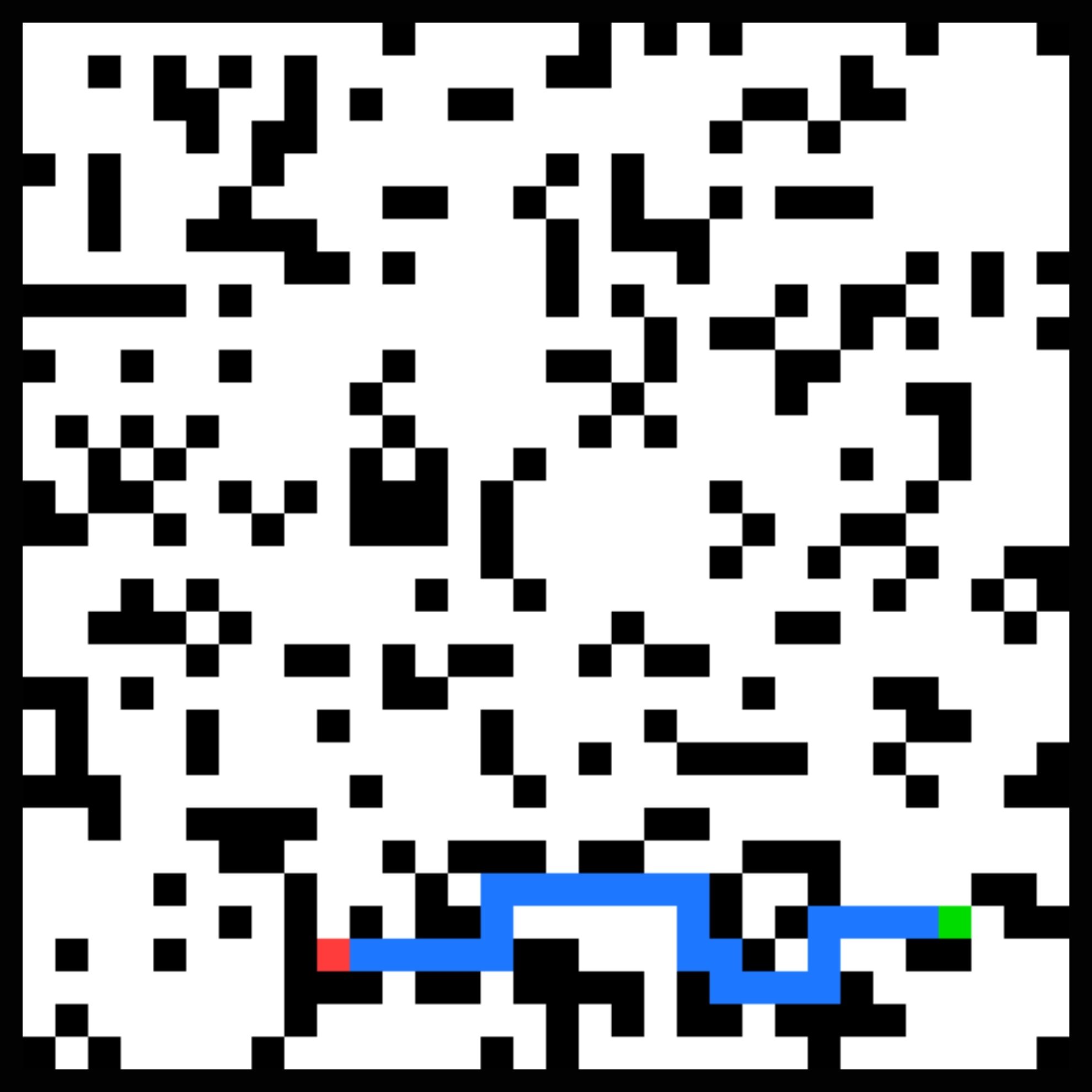} \\
\includegraphics[width=0.24\textwidth]{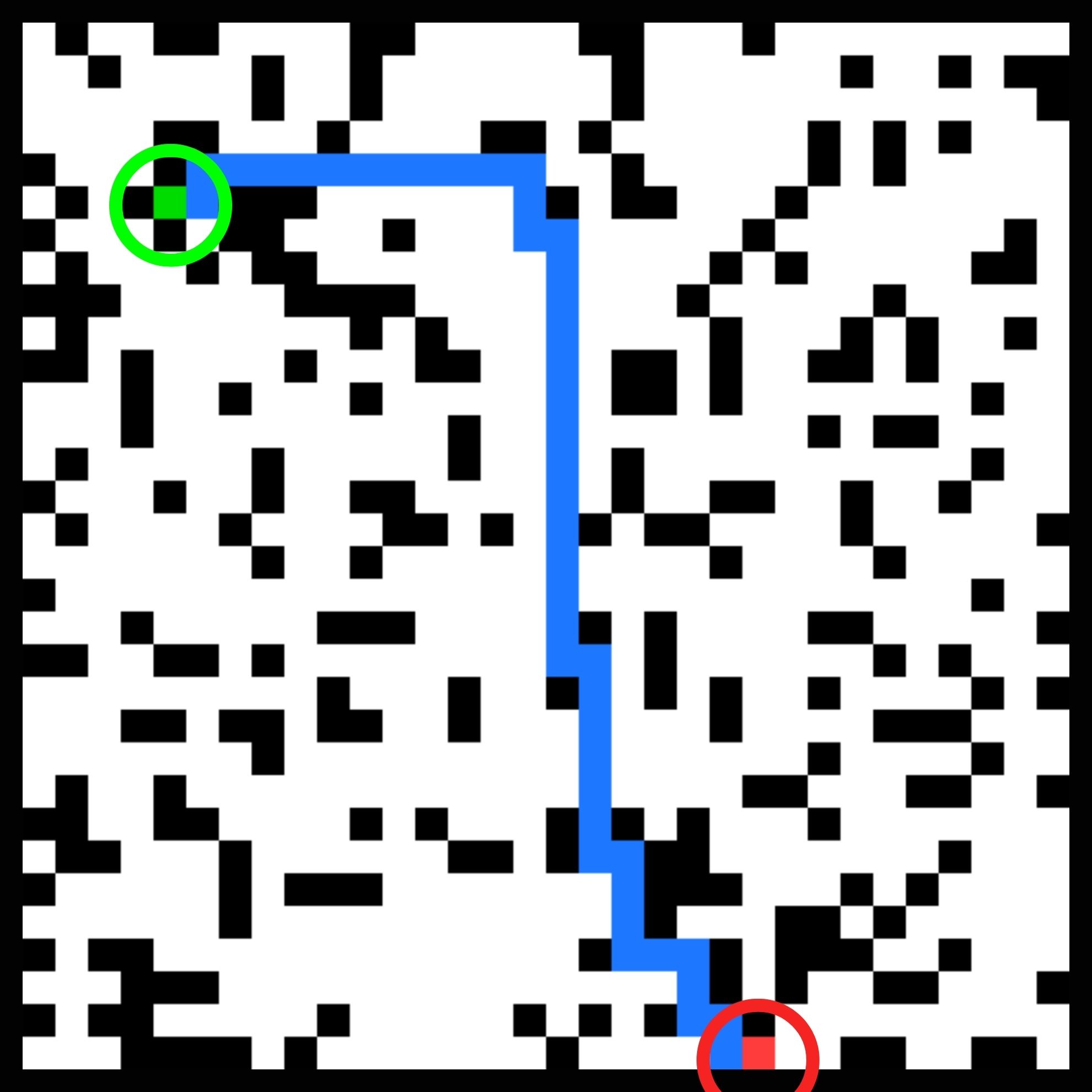} &
\includegraphics[width=0.24\textwidth]{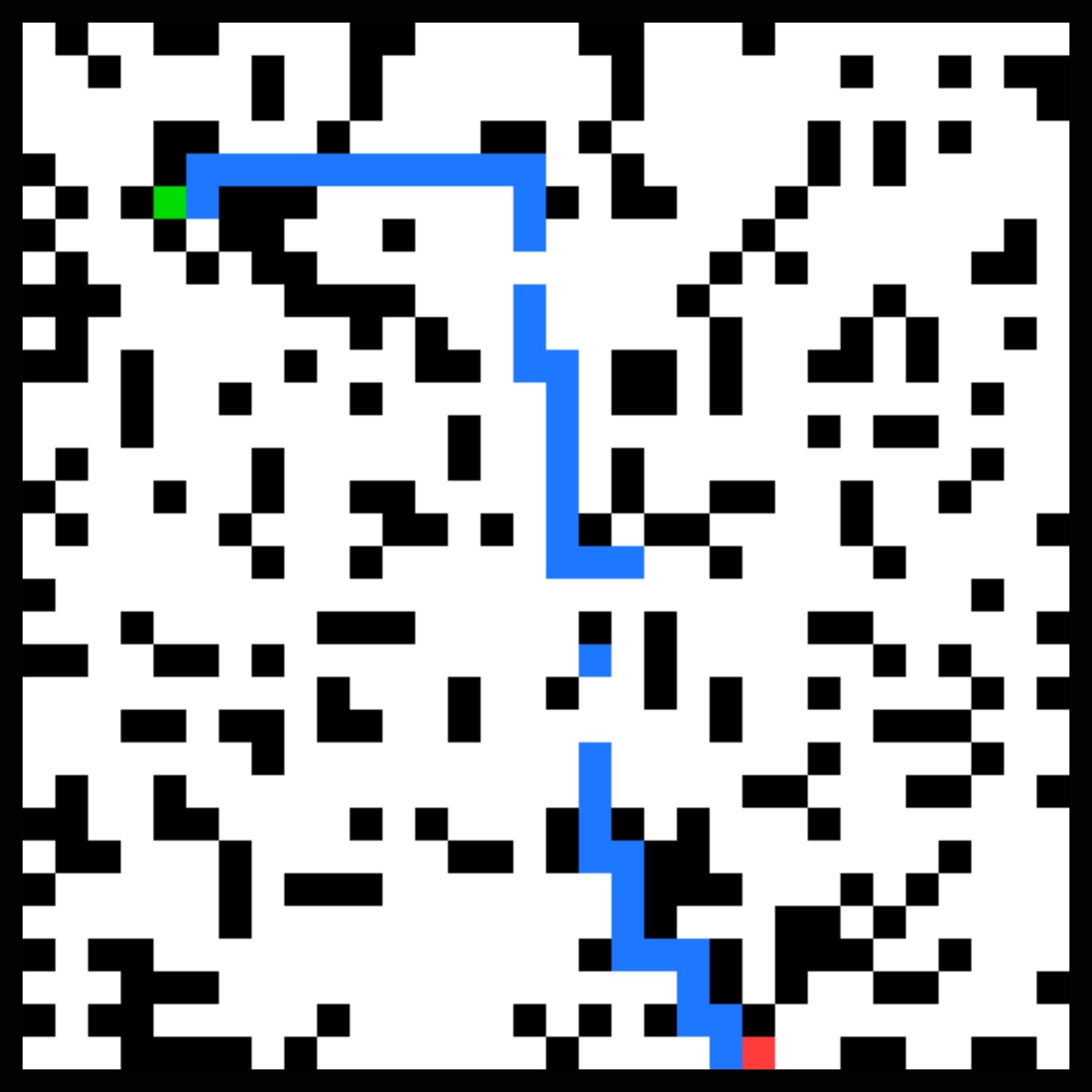} &
\includegraphics[width=0.24\textwidth]{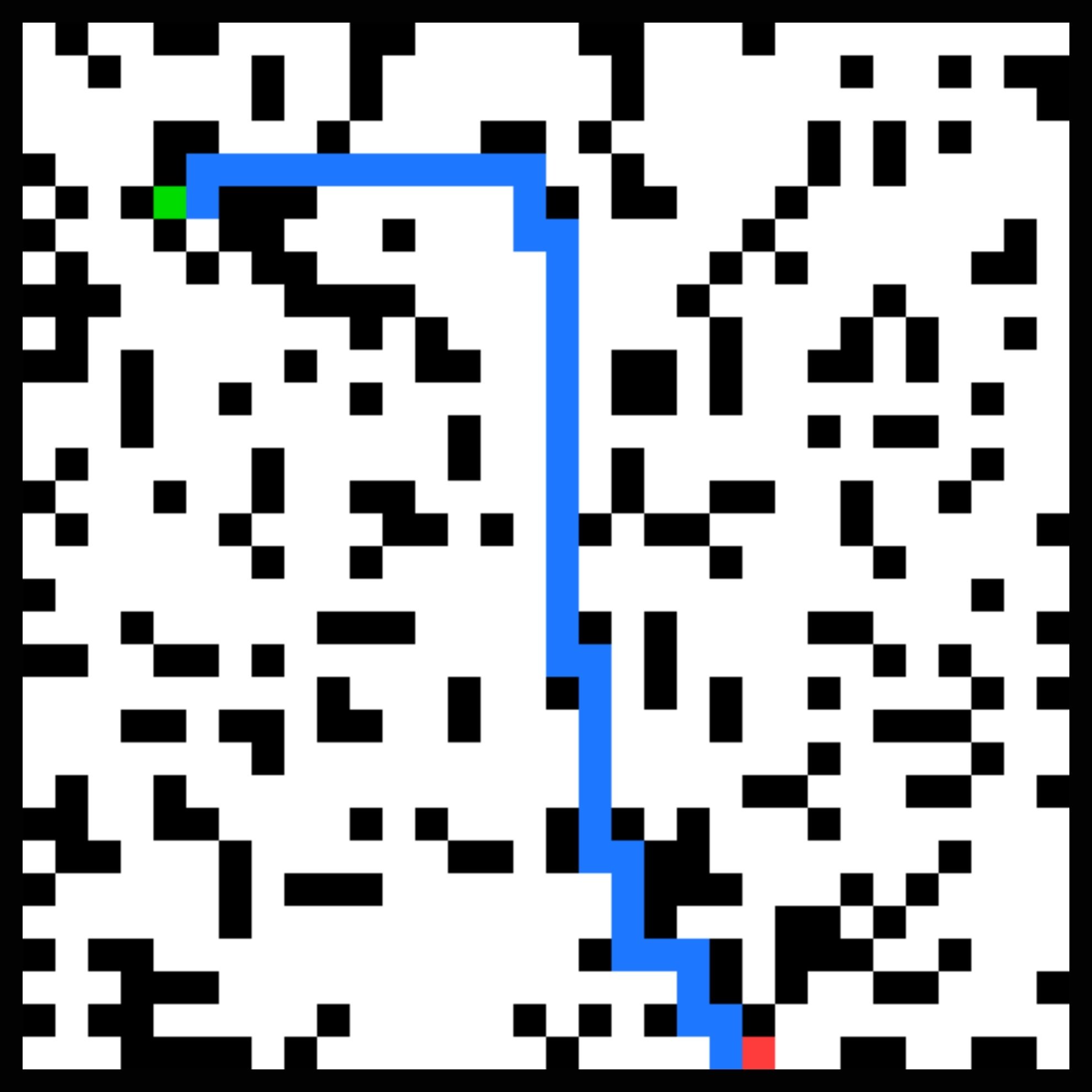} &
\includegraphics[width=0.24\textwidth]{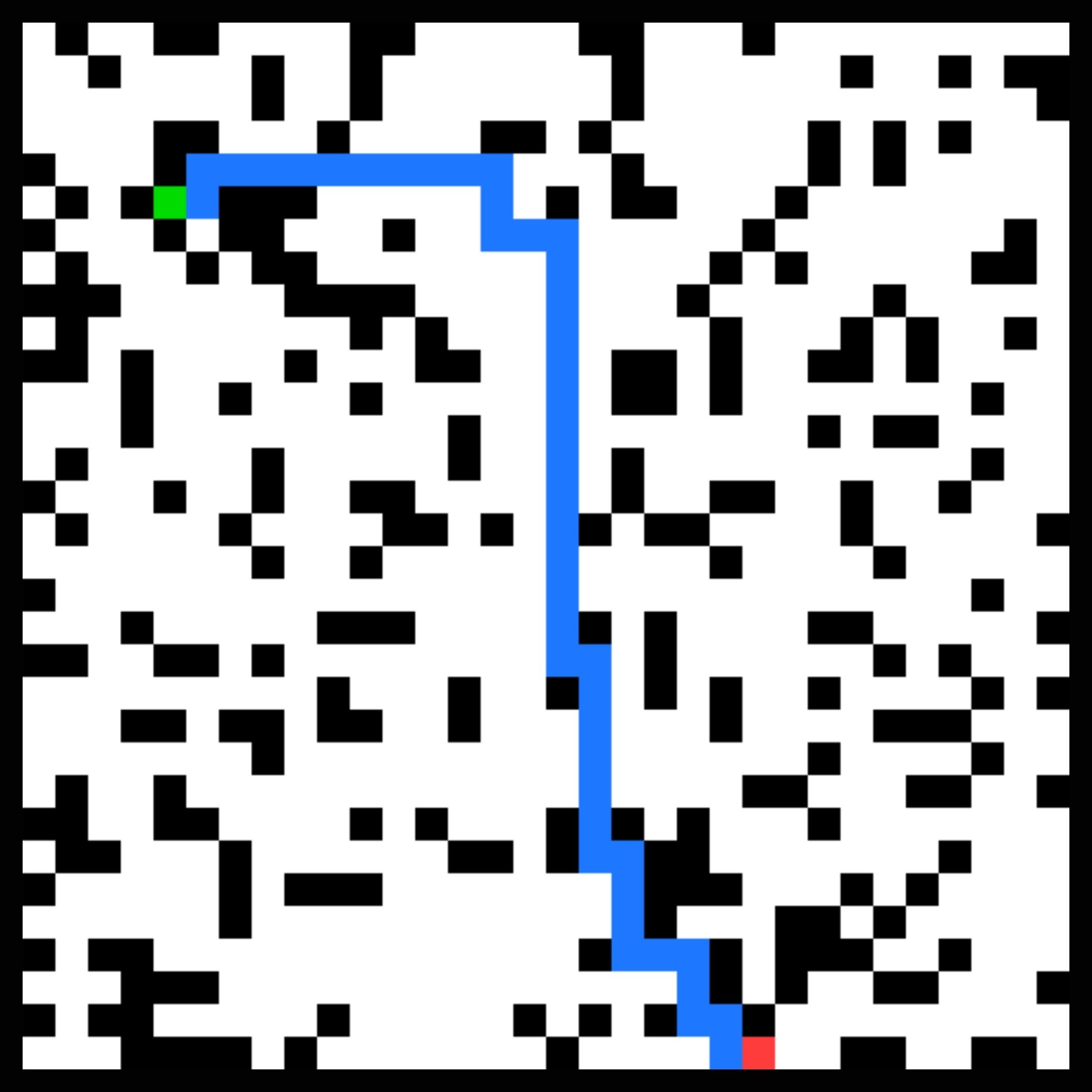} 
\end{tabular}
\caption{\textbf{Qualitative comparison of paths generated by CNN, DiffPlanner, and FlowPlanner on $32\times32$ mazes.}}
\label{fig:32x32}
\end{figure*}

We compare the quality of the path generated by three model variants: a baseline CNN (U-Net), DiffPlanner, and FlowPlanner, as shown in Figs. \ref{fig:48x48} and \ref{fig:32x32}. The baseline CNN directly predicts the path map based on a three-channel maze description (walls, start, goal), while DiffPlanner and FlowPlanner iteratively construct a solution, starting with random noise and gradually transforming it into a valid trajectory using a diffusion process and a trained vector field, respectively. The results in Table \ref{tab:cnn_diff_flow} demonstrate a clear advantage of generative methods (FlowPlanner and DiffPlanner) over the baseline model, and especially FlowPlanner consistently outperforms the other methods. DiffPlanner also achieves good performance results, but it performs slightly worse than FlowPlanner in metrics related to path structure. All models achieve \textit{Length Ratio} values close to one, indicating that the lengths of the generated paths are comparable to the optimal solutions identified by the A* algorithm. Notably, only the GenPlanner methods successfully combine this length property with a high degree of correctness and regularity in the trajectories. 
Both DiffPlanner and FlowPlanner more frequently generate valid start-to-goal connections, which is evident from the high \textit{Validity} values. These differences become more pronounced in larger mazes.  FlowPlanner excels in metrics related to the path structure, achieving the highest \textit{Single-Path} and the lowest \textit{Branch Rate}. This suggests that the generated trajectories are primarily single-path and free of branches.  

To investigate the impact of individual condition components on the quality of generated paths, we conducted an ablation study for mazes of size $48 \times48$. We analyzed variants in which the model received no condition, only the start and destination masks, only the obstacle map, and the full condition containing all three channels. The results are presented in Table \ref{tab: cond}. The summary shows that the absence of any of the condition channels leads to a drastic deterioration in the quality of generated trajectories. 

In particular, the model receiving only information about the start and destination points is unable to generate correct paths, resulting in a zero value of the \textit{Validity} metric. On the other hand, using only the obstacle map allows for partially correct solutions, but the generated paths are short and unstable. The best results are obtained for the full condition containing the obstacle map, start, and destination, confirming that all three channels are necessary for the correct path planning. 

The \textit{N/A} value of the \textit{Length Ratio} metric for the configuration without any conditioning results from the lack of valid trajectories generated (\textit{Validity}=0). This result confirms that effective planning requires explicit information about the environment's structure and endpoints. Configurations using only the start and target or only the obstacle map lead to significantly worse results. Example visualizations for this ablation are shown in Fig. \ref{fig: ablation}.

\begin{table}[!t]
 \centering
\caption{\textbf{Quantitative ablation results for FlowPlanner different conditioning channel combinations on $48 \times48$ mazes.} Each condition channel is crucial for accurately generating the trajectory.}
\setlength{\tabcolsep}{4.5pt}
{\fontsize{6.2pt}{6pt}\selectfont
\begin{tabular}{lcccccc}
\toprule
\multirow{2}{*}{Config} & \multicolumn{2}{c}{Conditioning} & \multicolumn{4}{c}{Metrics} \\ 
\cmidrule(lr){2-3} \cmidrule(lr){4-7}
 & Start,End & Walls & Validity (\%) $\uparrow$ & Single-Path (\%) $\uparrow$ & Length Ratio $\downarrow$ & Branch-Rate (\%) $\downarrow$ \\
\midrule
1    & \xmark & \xmark & 0.00 & 33.70 & N/A & 0.18 \\
2    & \cmark & \xmark & 6.40 & 6.30 & 1.06  & \textbf{0.03}  \\
3    & \xmark & \cmark & 0.10 & 73.30 & \textbf{1.00} & 0.22 \\
\midrule
Ours & \cmark & \cmark & \textbf{88.00} & \textbf{86.10} & 1.02 & 0.09 \\
\bottomrule
\end{tabular}

}
\label{tab: cond}
\end{table}

\begin{figure}[!t]
\centering
\setlength{\tabcolsep}{1pt}
\renewcommand{\arraystretch}{0.8}
\begin{tabular}{cccccc}
 & None & Start,End  & Walls & Start,End,Walls & Ground Truth \\
\rotatebox{90}{\hspace{6pt}{FlowPlanner}} &
\includegraphics[width=0.19\textwidth]{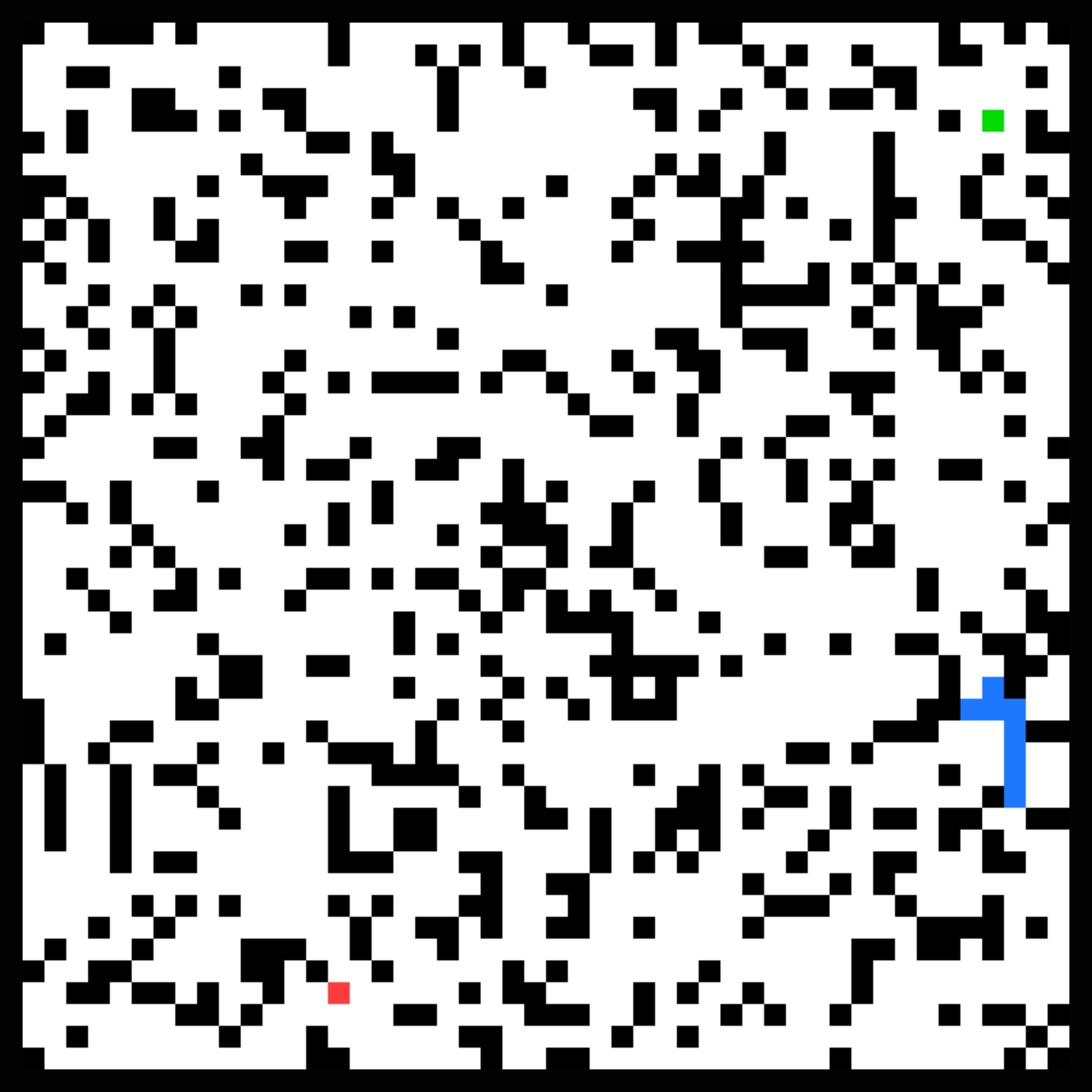} &
\includegraphics[width=0.19\textwidth]{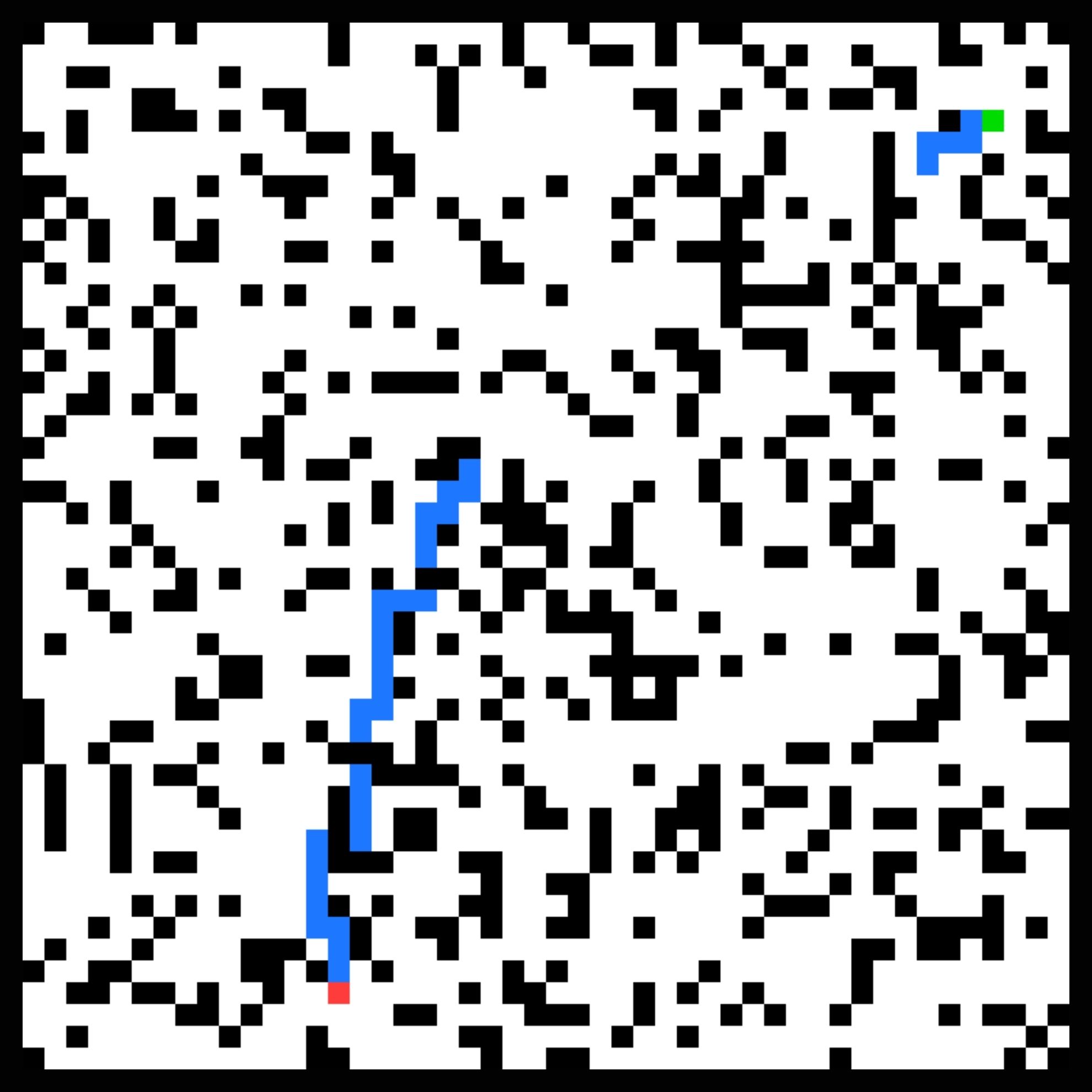} &
\includegraphics[width=0.19\textwidth]{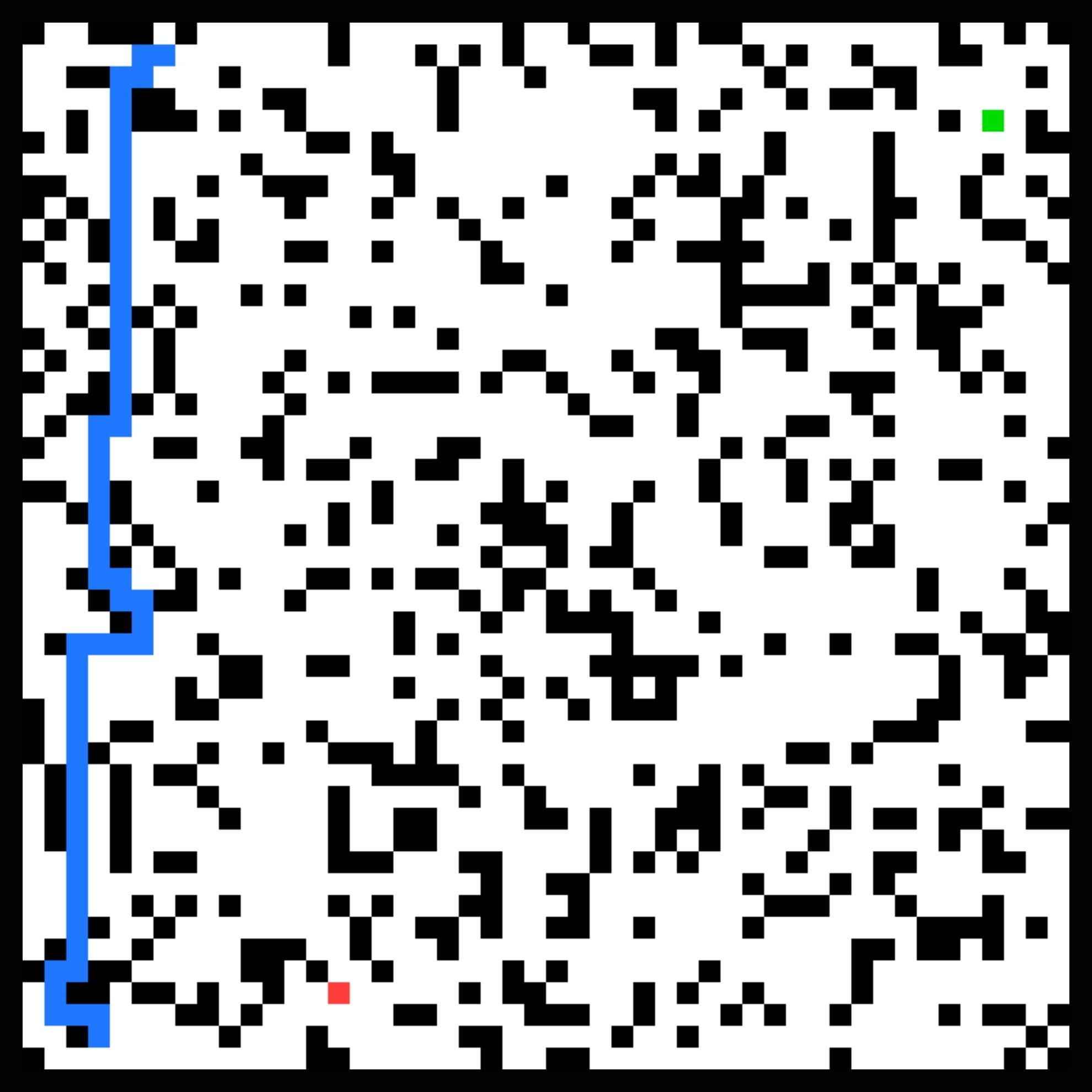} &
\includegraphics[width=0.19\textwidth]{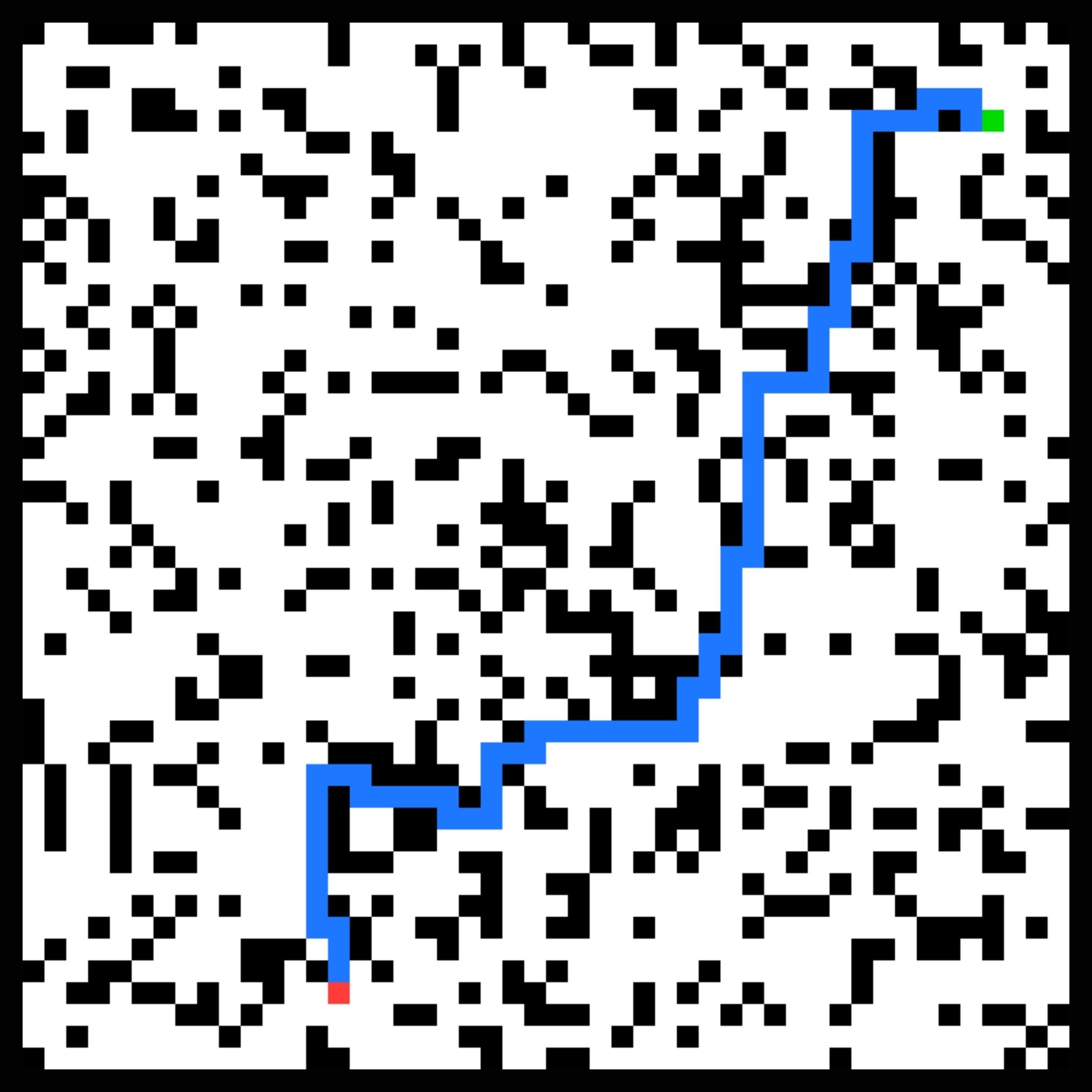} &
\includegraphics[width=0.19\textwidth]{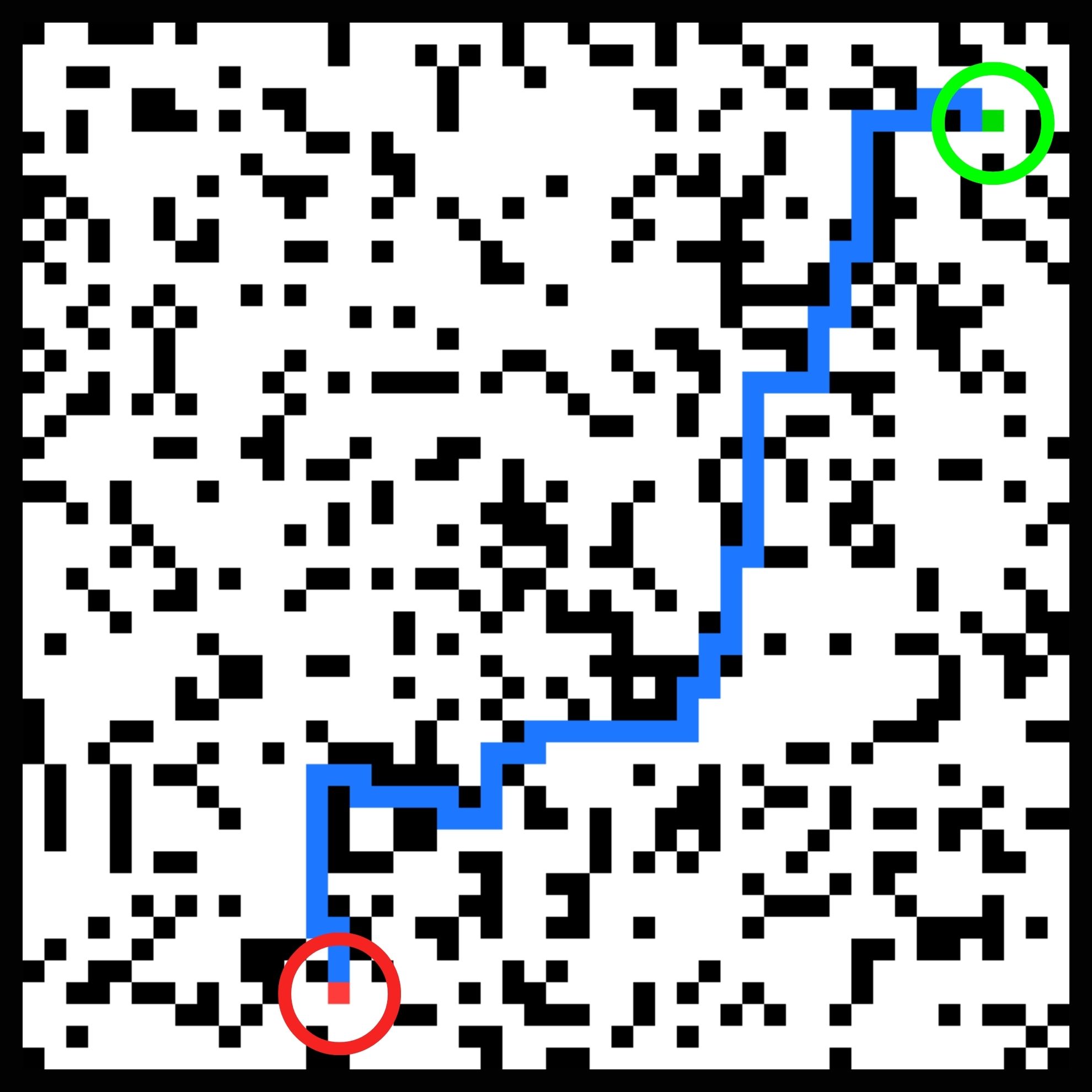} \\
\end{tabular}
\caption{\textbf{Qualitative comparison on FlowPlanner using different conditioning channel combinations on $48 \times48$ mazes.} For \textbf{Start, End} model knows where the path begins and ends, but the path is not valid. For \textbf{Walls}, the model doesn't know where to go but can take obstacles into account. \textbf{None } means without conditioning. For \textbf{Start, End, Walls}, the model achieves the best path.}
\label{fig: ablation}
\end{figure}

Table \ref{hyperparaemtrs_steps} shows the effect of the number of generation steps on the quality of the results for FlowPlanner and DiffPlanner. Reducing the number of steps leads to a gradual deterioration in quality in both cases, but this effect is significantly weaker for FlowPlanner. Even with 10 steps, FlowPlanner maintains high values for the \textit{Validity} and \textit{Single-Path} metrics, as well as a low \textit{Branch-Rate}, indicating the stability of the generation process even with a strong reduction in the number of iterations. In contrast, DiffPlanner is much more sensitive to reducing the number of steps. For a small number of iterations, we observe a sharp decrease in \textit{Single-Path} and strong increase in \textit{Branch-Rate}, indicating the formation of fragmented and branched trajectories. For just one step, DiffPlanner practically loses its ability to generate meaningful paths. These results indicate that the flow approach is more resistant to aggressive iteration limiting, enabling efficient generation while maintaining high quality. This represents a significant advantage of FlowPlanner over DiffPlanner in scenarios where speed is crucial.

\begin{table}[!t]
 \centering
\caption{\textbf{Impact of the number of sampling steps on the performance of FlowPlanner and DiffPlanner for $48\times48$ mazes.} FlowPlanner is characterized by high stability regardless of the number of generation steps.}
\setlength{\tabcolsep}{6pt}
{\fontsize{8pt}{8pt}\selectfont
\begin{tabular}{llcccc}
\toprule
Model & Steps $T$ & Validity $\uparrow$ & Single-Path $\uparrow$ & Length Ratio $\downarrow$ & Branch-Rate $\downarrow$ \\
\midrule

\multirow{6}{*}{FlowPlanner}
 & 50  & \textbf{88.00} & \textbf{86.10} & 1.02 & \textbf{0.09} \\
 & 30  & 87.70 & 85.50 & 1.02 & 0.10 \\
 & 20  & 86.40 & 84.60 & 1.02 & 0.10 \\
 & 10  & 85.80 & 83.00 & 1.02 & 0.15 \\
 & 5   & 79.10 & 75.90 & 1.01 & 0.14 \\
 & 1   & 23.70 & 20.40 & \textbf{1.00} & 0.65 \\
\midrule

\multirow{6}{*}{DiffPlanner}
 & 50  & 89.00 & \textbf{76.10} & 1.04 & \textbf{0.47} \\
 & 30  & 88.00 & 71.10 & 1.04 & 0.62 \\
 & 20  & 85.50 & 65.60 & 1.03 & 0.71 \\
 & 10  & 81.30 & 52.00 & 1.03 & 1.27 \\
 & 5   & 65.70 & 10.70 & 1.03 & 3.91 \\
 & 1   & \textbf{98.90} & 0.00 & \textbf{1.00} & 71.05 \\
\bottomrule
\end{tabular}
}
\label{hyperparaemtrs_steps}
\end{table}
\section{Conclusion}
We presented GenPlanner, an approach to path planning based on generative models, along with two variants: DiffPlanner and FlowPlanner. We showed that models using diffusion and flow matching can be treated not only as data generators but also as reasoning and planning mechanism. Experiments showed that the proposed methods significantly outperform the baseline CNN model in terms of the correctness of generated trajectories, their regularity, and structural stability. In particular, FlowPlanner achieves the best results in all key metrics, maintaining high quality even when the number of generation steps is severely limited.

%
%
\bibliographystyle{splncs04}
\bibliography{genplanner}
\end{document}